\documentclass{article}

\usepackage{subfigure}
\usepackage{url}
\usepackage{amsmath}
\usepackage{mathtools}
\usepackage[noend]{algorithmic}
\usepackage{algorithm}
\usepackage{enumerate}
\usepackage{rotating}
\usepackage[center]{caption}
\usepackage{tabularray}

\usepackage{multirow}
\usepackage{enumitem}
\usepackage{xcolor}
\usepackage{orcidlink}
\usepackage[inkscapeformat=png]{svg}

\usepackage{times}  % DO NOT CHANGE THIS
\usepackage{helvet}  % DO NOT CHANGE THIS
\usepackage{courier}  % DO NOT CHANGE THIS
\usepackage{graphicx} % DO NOT CHANGE THIS
\usepackage[numbers]{natbib}  % DO NOT CHANGE THIS AND DO NOT ADD ANY OPTIONS TO IT
\usepackage{caption}

\usepackage{arxiv}
\usepackage{amsmath}
\usepackage{graphicx}
\usepackage{xcolor}
\usepackage{amssymb}
\usepackage{yfonts}
\usepackage{hyperref}
\usepackage{textalpha}

\title{DeliverAI: A Distributed Path-Sharing Network based solution for the Last Mile Food Delivery Problem}

\author{
  Ashman Mehra\\
  Dept. of CSIS, BITS Pilani K K Birla Goa Campus, Goa, India\\
  \texttt{f20212508@goa.bits-pilani.ac.in}
\And
Snehanshu Saha\\
Dept. of CSIS and APPCAIR,  BITS Pilani Goa, India\\
HappyMonk AI, Bangalore, India\\
\texttt{snehanshus@goa.bits-pilani.ac.in}
\And
Vaskar Raychoudhury\\
Department of Computer Science and Software Engineering, Miami University, Miami, USA\\
\texttt{vaskar@gmail.com}
\And
Archana Mathur\\
Dept. of Information Science and Engineering, Nitte Meenakshi Institute of Technology, India\\
\texttt{mathurarchana77@gmail.com}
}

\begin{document}

\maketitle              % typeset the header of the contribution
\begin{abstract}
Delivery of items from the producer to the consumer has experienced significant growth over the past decade and has been greatly fueled by the recent pandemic. Amazon Fresh, Shopify, UberEats, InstaCart, and DoorDash are rapidly growing and are sharing the same business model of consumer items or food delivery. Existing food delivery methods are sub-optimal because each delivery is individually optimized to go directly from the producer to the consumer via the shortest time path. We observe a significant scope for reducing the costs associated with completing deliveries under the current model. We model our food delivery problem as a multi-objective optimization, where consumer satisfaction and delivery costs, both, need to be optimized. Taking inspiration from the success of ride-sharing in the taxi industry, we propose DeliverAI - a reinforcement learning-based path-sharing algorithm. Unlike previous attempts for path-sharing, DeliverAI can provide real-time, time-efficient decision-making using a Reinforcement learning-enabled agent system. Our novel agent interaction scheme leverages path-sharing among deliveries to reduce the total distance traveled while keeping the delivery completion time under check. We generate and test our methodology vigorously on a simulation setup using real data from the city of Chicago. Our results show that DeliverAI can reduce the delivery fleet size by 12\%, the distance traveled by 13\%, and achieve 50\% higher fleet utilization compared to the baselines.

\end{abstract}

\section{Introduction}
\label{sec:intro}
\subsection{Motivation}
Online Food Delivery (OFD) services are emerging as crucial components in the consumer market and are changing the way that goods are accessed and consumed \cite{CustomerTrends}. Last-mile delivery services help in bringing together producers and consumers. Consumers enjoy convenience, save time, get more accessibility, and leverage the advantages of a competitive market \cite{vakulenko2019service} while the producers get access to a larger and more diverse market with more opportunities to expand without worrying about attracting consumers. The success of last-mile delivery services in e-commerce has enabled companies like UberEats, Amazon Fresh, Doordash, and InstaCart to venture into the OFD industry in the past decade. The growth of last-mile delivery services has also recently been fueled by the outbreak of COVID-19 \cite{dsouza2021online}, when these services played an important role in delivering hygienic food to consumers and helping food outlets survive the pandemic. As a result of this, the global revenue of the OFD industry is rapidly rising from \$107.4 billion in 2019 to an expected \$182.3 billion by 2024 \cite{Statista}. These statistics show that even as the pandemic settles, the demand for these services is steadily increasing, and the OFD industry is here to stay.

Technological innovations in the last-mile delivery operations aim to revolutionize this industry by minimizing transport costs and increasing customer satisfaction (minimizing failures, delivery time, and other opportunity costs)  \cite{prasetyo2021factors,mangiaracina2019innovative}. One such work of interest lies in routing algorithms. These routing algorithms attempt to efficiently manage resources (fleet of vehicles, manpower, etc) and find routes (from source to destination) that minimize time and distance traveled for completing the deliveries. Traditional route optimization algorithms like Dijkstra's Shortest Path Algorithm \cite{dijkstra2022note} and Floyd Warshall's Algorithm for the shortest path between all node pairs \cite{warshall1962theorem} were the first ones to solve the routing problem. However, these algorithms optimize each source-destination pair individually and fail to provide any additional advantage when many deliveries need to be delivered simultaneously. They also do not provide any efficient mechanism to reallocate resources like vehicles. 

To handle these limitations, a more generalized formulation called the Vehicle Routing Problem (VRP - a modified version of the Travelling Salesman Problem) came into the picture\cite{pillac2013review}. The solutions to VRP provide three major benefits - the ability to serve many customers together, route a vehicle to complete multiple deliveries simultaneously and handle real-time requests. However, delivery problems are also plagued with dynamic conditions in the underlying network graph - traffic conditions in any city vary with the time, day, month, etc. The above-mentioned algorithms do not have any capacity to learn the traffic patterns in the city and make intelligent decisions based on them.  They are also computationally expensive (as VRP is an NP-hard problem). Hence, they often fail to cope with dynamically changing traffic unless they rerun from scratch and incur delivery delays. More importantly, these deterministic algorithms are more suited to a single objective - minimizing time or distance using fixed resources (vehicles). In reality, however, there are multiple objectives to be balanced, as discussed in Section \ref{sec:objectives}. 

Artificially intelligent algorithms like \textit{Reinforcement Learning} can solve these problems. Reinforcement Learning (RL) \cite{sutton} is a machine learning approach wherein an agent attempts to explore the environment and learn an optimal action policy from its experiences. This action policy helps the agent to take the best possible action in order to achieve the goal of the problem under consideration. In a typical reward model, the agent is provided with the current state of the environment to take an action. The action is then executed in the environment, and the agent receives a reward for the action it took. The reward scheme is representative of the overall objective that is to be achieved by the agent. If the agent takes a favorable action, it is provided with a positive reward,  and if the agent takes a non-favorable action, then it is provided with a possibly negative reward (penalty). The agent updates its action policy based on the reward it receives. This process of taking an action and getting the observation is repeated till the agent converges to an optimal action policy. Using Reinforcement learning helps DeliverAI to learn the underlying traffic network iteratively and adapt to the changing conditions while retaining past experiences.

In this paper, we present DeliverAI (pronounced as deliver-eye) - a multi-agent-based system that can intelligently (and greedily) assign multi-hop routes to deliveries using \textit{Reinforcement Learning}. With DeliverAI, we introduce the novel approach of \textit{path-sharing}, wherein deliveries going in the same direction can combine and travel together. We take this inspiration from the success of ride-sharing in the taxi industry, which recently became very popular \cite{StatistaTaxi} because sharing minimizes operational costs. Similar to passengers sharing a taxi, we devise a mechanism for food deliveries to share delivery vehicles for a part of their journey. Unlike ride-sharing in taxis, in path-sharing, food deliveries need not travel in the same vehicle but rather change vehicles in their journey and share routes with multiple deliveries. To enable this mechanism, we first set up a delivery network on the city map that the deliveries traverse to reach their final destination. Then, we provide a multi-agent setup wherein the RL agents interact with each other and determine the routes of the deliveries. Using the delivery network and the multi-agent communication, DeliverAI plans multi-hop and path-sharing routes for the deliveries. Finally, we test DeliverAI on real data from the city of Chicago and provide empirical evidence to demonstrate its performance. 

\subsection{Contributions}
The key contributions of the paper are listed as follows:
\begin{itemize}
    \item We introduce the first-ever path-sharing food delivery network for deliveries using Reinforcement Learning to make intelligent, real-time, and dynamic decisions. We provide a novel method to use the Q-values of Reinforcement Learning Agents for inter-agent communication in a greedy setup.
    \item We model the problem as a Markov Decision Process (MDP) with Multi-Objective Optimization and define key performance metrics to analyze the performance of DeliverAI. We have built a large-scale producer-consumer food delivery data set for the city of Chicago using open-source databases and tested DeliverAI vigorously using these data sets.
    \item Our in-depth experimentation shows that the path-sharing introduced by DeliverAI can achieve significant benefits for delivery companies by reducing the distance traveled by 13\%, reducing the fleet size by 12\%, and providing 50\% better fleet management and utilization when compared to the baselines.
\end{itemize}

\section{Related work}

Due to rising demand for more agile, cost-effective, and eco-friendly delivery methods utilizing autonomous vehicles, drones, and last-mile solutions, researchers have turned their attention to shared-economy models. These models are often based on ride-sharing principles - sharing resources like vehicles among passengers or goods. In the literature review, we have identified various solutions that use a multi-hop approach to solve the delivery problems and divided them into two categories - (i) Solutions using deterministic and heuristic algorithms and (ii) Reinforcement Learning based algorithms. 

Various papers have attempted to provide multi-hop solutions for parcel delivery using deterministic or heuristic algorithms. CrowdDeliver \cite{crowddeliver} is one of the first papers to explore multi-hop route planning for package deliveries using the existing passenger-taxi network. Their 2 phase approach deals with finding the shortest delivery paths using historical data and then using AdaPlan to find the routes dynamically (at run-time). PPtaxi \cite{pptaxi} also presents a Joint Transportation problem - Non-Stop Package Delivery (NPD), where they aim to utilize the empty cargo space in passenger-carrying vehicles for complete package delivery. The approach involves predicting passenger flows and then using a Dijkstra-based optimal route planning algorithm (DOP) for planning delivery routes. \cite{mdmpmp} solves their Multi-Driver Multi-Parcel Matching Problem (MDMPMP) using  a similar approach - passenger vehicles take a detour from their original planned path to pick up and drop off delivery parcels. They use a Time Expanded Graph (TEG) based heuristic to sync the path of the delivery parcel with the arrival of vehicles at various nodes using the A* algorithm. While the algorithms in \cite{crowddeliver,pptaxi,mdmpmp} are effective for their problems, they are passenger-centric, i.e. the passengers are given superior priority, and hence the routes planned for the deliveries are sub-optimal and time inefficient. Additionally, their multi-hop strategies deal with very little planning for the vehicle requirement and their routes (which are largely pre-determined and fixed).

As mentioned in Section \ref{sec:intro}, Reinforcement Learning is adept at solving multi-objective problems, and, in recent years, delivery routing algorithms using RL have become popular. \cite{fleetmanagement} uses Contextual Multi-Agent Reinforcement Learning (MARL) to solve the Fleet Management Problem to re-allocate vehicles and cater to deliveries. However, in their setup, ride-sharing is possible, with each delivery traveling in one vehicle throughout (i.e., deliveries cannot change vehicles in their journey). FlexPool \cite{flexpool} provides the first RL-enabled joint transport system where delivery packets can change vehicles at hop zones (which we call hotspots). They model the vehicles as agents and use a Double Deep Q-Learning Network (DDQN) to train them. They further build upon their work in PassGoodPool \cite{passgoodpool} to include pricing into their algorithm. The RL algorithms talked about so far again deal with a joint transportation system where passengers are given more importance. To the best of our knowledge, DeepFreight \cite{deepfreight}  is the first to provide an RL-based multiple transfer approach to solve the Delivery Transport Problem alone. They use a MARL framework called Q-MIX to train their agents (vehicles). Since they deal with freight transportation, their reward model focuses more on planning the path of the vehicle and is less focused on the time taken to complete the delivery. we identify some key differences between our approach and the ones in the literature. Firstly, food deliveries are time-sensitive and need to be delivered quickly and safely. Hence our focus is on ride-sharing as well as keeping the delivery completion time in the acceptable range. Secondly, we provide a novel approach that is delivery-centric i.e. rather than formulating the solution as routing a vehicle, we consider routing the delivery through a fixed network (which we call Overlay Network, Section \ref{sec:delivery-network}). Unlike the previous attempts which use a fixed fleet size, DeliverAI incorporates the fleet requirement calculation in its algorithm and attempts to reduce the fleet size as well.

\renewcommand{\arraystretch}{1.3}
\begin{table}[t]
\centering
\caption{Variables Used}
\centering
\setlength\tabcolsep{2.0pt}
\begin{tabular}{|l|l|}
\hline
VARIABLE & MEANING                                                                                                                                                          \\ 
\hline
\hline
$D$               & Set of Deliveries,~$d_i \in D$                                                                                                                                            \\ 
\hline
$H$               & Set of Hotspots in the Overlay Network (ON),~$h_i \in H$                                                                                                                                       \\ 
\hline
$V$               & Set of Vehicles,~$v_{i} \in V$                                                                                                                                            \\ 
\hline
$d_i^{src}$         & Source hotspot of $d_i$, $d_i^{src} \in H$                                                                                                                               \\ 
\hline
$d_i^{dest}$        & Destination hotspot of $d_i$, $d_i^{dest} \in H$                                                                                                                            \\ 
\hline
$d_i^{start}$       & Start time of $d_i$                                                                                                                                                       \\ 
\hline
$d_i^{end}$        & End time of $d_i$                                                                                                                                                         \\ 
\hline
$d_i^{path}$       & \begin{tabular}[c]{@{}l@{}} Ordered tuple of the hotspots visited by $d_{i}$ in its path, \\ $d_{i}^{path} = (x_1, x_2, .... , x_n), x_i \in H$, $d_i^{src} \notin d_i^{path}$\end{tabular} \\ 
\hline
$d_i^{next}(t)$        &  the hotspot that $d_i$ is going to (at time $t$)                                                                                                                                   \\ 
\hline
$d_i^{curr}(t)$        &  the hotspot that $d_i$ is situated (at time $t$)                                                                                                                                   \\ 
\hline
$d_i^{wait}(t)$        &  total waiting time $d_i$ (at hotspots)                                                                                                                                   \\ 
\hline
$d_i^{share}(t)$       & $\begin{cases} \text{$NULL$} & \text{if $d_i$ is not sharing at $t$} \\ \text{$d_j$  $(\in D)$} & \text{if $d_i$ and $d_j$ are sharing at $t$} \\ \end{cases}$      \\ 
\hline
$h_i^{veh}(t)$      & Number of vehicles in $h_i$ at time $t$                                                                                                                                        \\ 
\hline
$h_i^{act}(t)$      & Number of active vehicles in $h_i$ at time $t$                                                                                                                                        \\ 
\hline
$h_i^{del}(t)$      & Number of deliveries at $h_i$ at time $t$                                                                                                                                        \\ 
\hline
$v_i^{dist}$      & Total Distance covered by vehicle ~$v_{i}$                                                                                                                                         \\ 
\hline
$v_i^{cap}$       & Capacity (number of deliveries) of vehicle ~$v_{i}$                                                                                                                                        \\ 
\hline
$agent_i$         & Agent trained for hotspot $h_i$                                                                                                                                         \\ 
\hline
$\pi$             & Joint Policy of all agents                                                                                                                                                \\ 
\hline
$A$               & Joint action space of all agents                                                                                                                                          \\ 
\hline
$S$               & Joint State space of all agents                                                                                                                                           \\ 
\hline
$Q_i$             & Action Value Function of $agent_i$                                                                                                                                        \\ 
\hline
$r_i$             & Reward function of ~$agent_i$                                                                                                                                                       \\ 
\hline
$T$               & Temperature parameter in~ $\pi_{boltzmann}$, $T > 0$                                                                                                                                                \\ 
\hline
$\epsilon$        & Epsilon parameter in $\pi_{epsilon}$ ,~$0 \le \epsilon \le 1$                                                                                                                             \\ 
\hline
$\alpha$          & Learning Rate in Q-learning~                                                                                                                                                         \\ 
\hline
$\gamma$          & Discount factor in Q-learning                                                                                                                                                       \\ 
\hline
$R_{agent}$       & Agent Interaction Range                                                                                                                                                   \\ 
\hline
$PAS_i(h)$           & Preferred Action Set for~$d_{i}$ when situated at hotspot $h$                                                                                                                         \\
\hline
$l(t)$               & \begin{tabular}[c]{@{}l@{}} Delivery load at time $t$ - number of deliveries \\ every minute\end{tabular}                                                                                                                            \\

\hline
$\tau$            & Delivery time window for delivery to be successful                                                                                                                             \\ 
\hline
$T_{sim}$         & Simulation duration                                                                                                                             \\ 
\hline
\end{tabular}
\end{table} 

\section{Problem Definition and Objectives}
\label{sec:problem-definition}
This section gives a detailed description of the delivery problem. We first define some key terms and definitions that will be used in this paper. Then we define the main objectives of our solution and the motivations behind choosing them.

\subsection{Problem Definition and Key Terms}

\begin{enumerate}[leftmargin=0pt]
    \item \textit{Producers and Consumers}: A Producer is an entity that produces food. These typically include restaurants, fast food centers, cafes, etc. A Consumer is an entity that places a food order with a producer and expects the food to be delivered at the desired location. This location could be apartments, houses, offices, etc.
    \item \textit{Delivery company}: A Delivery company is an entity that arranges for the food order to be picked up from the producer and delivered to the consumer. In the OFD industry, the producer only prepares the food, and the delivery of the food is completely handled by the delivery company.
    \item \textit{Delivery (food order)}: A Delivery is an abstraction of the food that has to be delivered to the consumer. Every delivery can be identified by the triplet $\langle producer, consumer, start-time \rangle$. The Delivery company has complete information about the locations of the producer (who will prepare it), the consumer (who will receive it), and the time it needs to be picked up from the consumer. A delivery is \textit{active} when it has been picked up from the producer but is yet to be completed (i.e. handed over to the consumer).
    \item \textit{Delivery Vehicle}: A Delivery Vehicle is a motor-operated vehicle that is used for completing deliveries. A Delivery Vehicle is said to be \textit{active} when it is carrying at least one delivery. Its speed remains constant throughout its operation.
    \item \textit{Path Sharing}: Deliveries traveling in the same delivery vehicle are said to be \textit{sharing}. Our problem deals with finding deliveries that can share a delivery vehicle for a part of their journey. Since the deliveries might need to ultimately reach different locations, they eventually stop sharing the delivery vehicle. This path-sharing, however, is not limited to a single continuous phase. Rather, it can be characterized by intermittent phases of sharing and non-sharing throughout the complete journey of the delivery.
    \item \textit{Delivery Hop}: A Delivery Hop (or \textit{hop}) is a discrete and uninterrupted movement of a delivery from one location to another. For instance, a delivery that is taken directly from the producer to the consumer is completed in one hop. This hop results in the handover of the delivery either directly to the end consumer or to the next delivery vehicle in the distribution chain.
    \item \textit{Agent}: An agent is an abstract entity with the primary responsibility of navigating deliveries to their destination. Each agent is trained to route deliveries to a particular destination. Agents also have the capacity to communicate with each other to facilitate sharing amongst deliveries. The details of this will be elaborated in further sections.
\end{enumerate}

\subsection{Multi-Objective Formulation }
\label{sec:objectives}
Routing algorithms are purposefully engineered with predefined objectives. Our modeling of the delivery routing problem comes under the domain of \textit{multi-objective optimization}.  We have carefully designed 3 essential objectives relating to minimizing the distance traveled, the delivery time for completing the deliveries, and the fleet (vehicle) requirement. We consider these as key objectives for the success of DeliverAI as they cater to the needs of the consumer as well as the goals of the delivery company. The details of our objectives are:
\vspace{5pt}
	\begin{enumerate}
    \item \textit{Minimizing Total Distance Travelled:} Path-sharing allows deliveries to travel together and utilize the same vehicle to reduce the distance traveled. This is extremely beneficial for the company to reduce fuel costs and the resultant environmental impact.
\begin{equation}
    \min ( \sum_{i} v_{i}^{dist} )
\end{equation}
\item \textit{Minimizing the Number of Delivery Vehicles:} One of the major operational costs lies in building a network of delivery vehicles and hiring drivers to operate these vehicles. With path-sharing strategies, our objective is to reduce the demand for delivery vehicles by allowing a vehicle to carry complete multiple deliveries concurrently.
\begin{equation}
    \min ( |V| ) \hspace{10pt}  \mbox{where  } |V| = \sum_{i} {v_i}
\end{equation}
    \item \textit{Minimizing Delivery Completion Time:} Path-sharing introduces synchronization delays wherein deliveries wait for each other (to meet at a common location) and take detours from their original (shortest) routes. While sharing definitely increases delivery time, DeliverAI aims to keep the additional time incurred to a minimum.
\begin{equation}
    \min (\sum_{i}(d_{i}^{end} - d_{i}^{start}))
\end{equation}
\end{enumerate}

The above objective functions are subject to the following constraints:
    \begin{equation}\label{eq:vehicle-constraint}
            v_i^{cap} \le 2, \hspace{10pt} v_i^{dist} > 0, \hspace{10pt} \forall v_i
    \end{equation}
    \begin{equation} \label{eq:hotspot-constraint}
        h_i^{veh} \ge h_i^{del}
    \end{equation}
    \begin{equation} \label{eq:time-constraint}
            0 \le d_i^{start} \le T_{sim},
            \hspace{10pt} d_i^{end} > 0 \hspace{10pt} \forall d_i
    \end{equation}

Equation \ref{eq:hotspot-constraint} ensures that in the case when no sharing can occur, each delivery is guaranteed dedicated access to a delivery vehicle for its transportation. Additionally, we assume that there is no delay caused in handing over the delivery between the vehicles. As in a typical multi-objective optimization, our objectives are also conflicting in nature. Using path-sharing reduces the distance traveled at the cost of increased delivery completion time. With DeliverAI, we aim to balance this trade-off and find a \textit{Pareto solution} to the problem where we choose paths for the deliveries that minimize the total distance traveled and the vehicle requirement by minimal degradation in delivery completion time. 

\section{Delivery Network Model}

\label{sec:delivery-network}
\begin{figure}
    \centering
    \includegraphics[width=\linewidth]{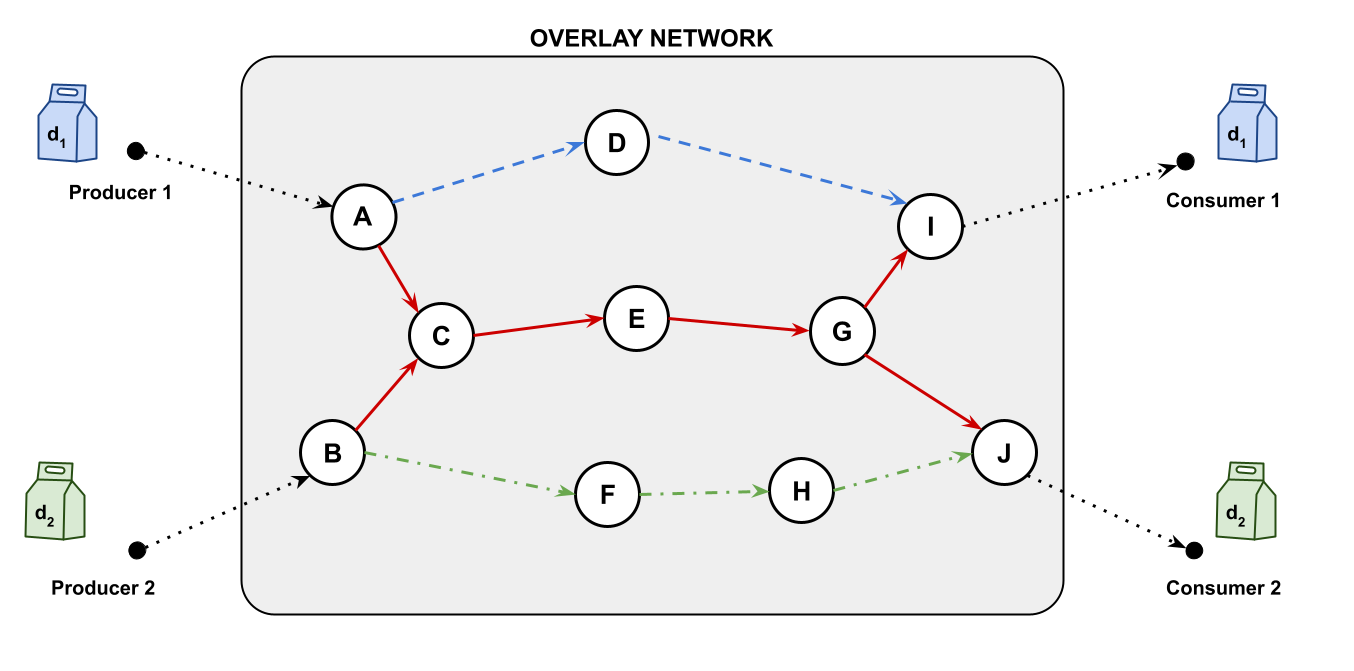}
    \caption{A schematic to show the path-sharing in a network of nodes (hotspots). In the network, the deliveries $d_1$ and $d_2$ can deviate from their shortest paths (shown by blue and green dotted arrows) to follow a common path (shown by red solid arrows) to save the total distance at the cost of the extra time taken to reach their destination.}
    \label{fig:path-sharing}
\end{figure}

Figure \ref{fig:path-sharing} depicts a situation with two deliveries - $d_1$ and $d_2$, starting at locations $A$ and $B$ and concluding at destinations $I$ and $J$, respectively. Consider the scenario where the individual paths of the deliveries are optimized to take the least time - $d_1$ follows the path $A \rightarrow D \rightarrow I$ and $d_2$ follows the path  $B \rightarrow F \rightarrow H \rightarrow J$. In contrast to this, consider that $d_1$ and $d_2$ share a part of their journey (as shown by the red path). They synchronize their arrival at $C$, share the common path $C \rightarrow E \rightarrow G$, and part ways at $G$ to travel to their final destinations individually. The cumulative effect of path-sharing is that the total distance traveled by $d_1$ and $d_2$ is reduced in comparison to the former scenario.

As shown by Figure \ref{fig:path-sharing}, there is a need to construct a graph network to enable the deliveries to share a common route. This \textbf{Overlay Network (ON)} is a fully connected graph (clique) with the nodes placed across the city (shown as $A,B,C...$).  We place a single dedicated delivery handler or router, known as \textbf{Hotspot}, at every node location. Hotspots are storage hubs where delivery vehicles can dock and exchange deliveries or store deliveries for a short duration. Delivery vehicles can hop from hotspot to hotspot while carrying deliveries to their final destination. Figure \ref{fig:overlay-network} shows a layer-by-layer construction of the Overlay Network.

Every delivery $d_i$ first reaches the hotspot closest to the producer ($d_i^{src}$) with the help of a \textbf{Peripheral Delivery Vehicle} (PDV). After entering the ON at $d_i^{src}$, the delivery hops from hotspot to hotspot till it reaches the hotspot closest to the consumer ($d_i^{dest}$) using \textbf{Core Delivery Vehicles} (CDV). This routing of $d_i$ through the ON is managed by $agent_{d_i^{dest}}$, which is specifically trained to navigate deliveries to $d_i^{dest}$. DeliverAI dynamically plans each hop of $d_i$ (the detailed procedure is described in Section \ref{sec:deliverai}). Once $d_i$ reaches $d_i^{dest}$, it is handed over to another PDV, which completes the last mile and takes the delivery to the consumer. Hence every delivery enters and exits the ON using PDVs and traverses the ON using CDVs.

The placement of the hotspots is crucial for the success of DeliverAI. To place these hotspots in the city of Chicago, we leverage the framework of \textit{census tracts}. Chicago is divided into smaller administrative units called census tracts \cite{CensusInfo}. We deploy one (and only one) hotspot in each of these census tracts. Since the hotspots are transit locations for the delivery, we wish to place them in busy areas of the city. Hence we choose to place them optimally at the centroid (by latitude and longitude) of all consumer locations within the census tract. Since each census tract has at most 8,000 people or 3,200 housing units (2020 census data \cite{CensusInfo}), selecting a single hotspot within each census tract offers the advantage of distributing the delivery workload evenly among these hotspots.

\section{Reinforcement Learning for Training RL Agents}
\label{sec:training}
The agents are a crucial part of DeliverAI as they manage the navigation through the Overlay network. The agents undergo a training process where they learn the paths connecting the various hotspots. Each agent is dedicated to learning the navigation to a unique hotpot. Formally, $agent_i$ learns paths from any hotspot $h_j$ $(j \ne i)$ to $h_i$. Once trained, $agent_i$ will be capable of routing all deliveries ($d_i$) with $d_{i}^{dest} = h_i$. In our RL setup, all agents train in the same environment (ON). Hence they share a common state space and action space. However, since each agent learns navigation to a different hotspot, each of them has a unique reward function. A formal definition of the setup is given:

\begin{figure}[h]
    \centering    \includegraphics[width=\linewidth]{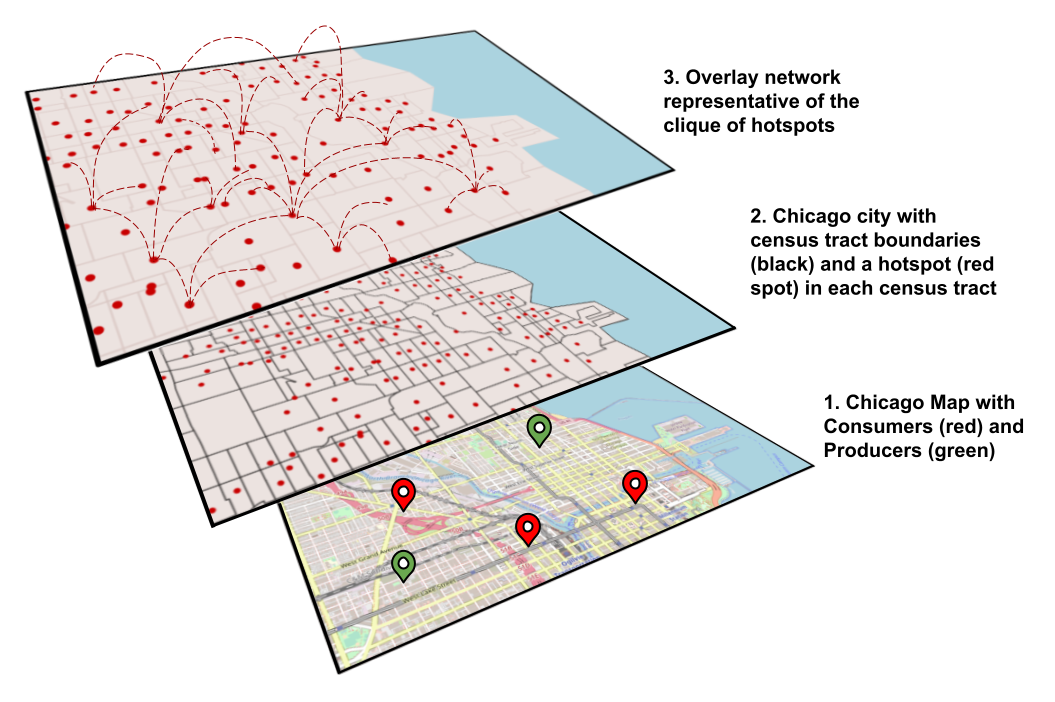}
    \caption{Figure shows the Overlay Network Visualisation on the map of Chicago. Layer 1 shows the map of Chicago with some consumer and producer locations (for representation). Layer 2 shows the division of the city into census tracts, with a hotspot placed in each tract. Layer 3 shows the Overlay Network clique, representing the network of delivery vehicles spanning across the city.}
    \label{fig:overlay-network}
\end{figure}

\begin{enumerate}
    \item \textit{(Joint) State Space} ($S$) - The state of an agent at any step (denoted by $s_t$) is its hotspot location in the Overlay network. The state space ($S$) comprises all the hotspots in the city i.e. $S = H$.
    
    \item \textit{(Joint) Action Space} ($A$) - The action taken by the agent in a step (denoted by $a_t$) is the hotspot it will hop to in the next step. Since the ON is a clique, we provide the agent with complete freedom to choose any hotspot from the ON as the action. Hence, like the state space, the action space ($A$) also comprises all the hotspots in the city i.e. $A = H$.
    
    \item \textit{Reward Function} - The reward function is a crucial component of any RL setup as it represents the objectives of the agent. Hence a high reward is guaranteed for reaching the destination, while a penalty (negative reward) is given for the time taken to complete every step i.e. the time taken to hop from the current hotspot to the next. In order to motivate the agent to explore the environment, a high penalty is also given if the agent decides to stay at the same hotspot. The reward function for $agent_i$ at ant time step $t$  is defined as:
    \[
        r_i(a_t|s_t) = \begin{cases}
      -1 & \mbox{if } a_t = s_t\\
      -t_{a_{t}s_{t}}^{norm} & \mbox{if } a_t \ne s_t \mbox{ and } a_t \ne h_{i}\\
      -t_{a_{t}s_{t}}^{norm} + 1 & \mbox{if } a_t \ne s_t \mbox{ and } a_t = h_{i}
      \end{cases}
    \]
Here the travel time between two hotspots, $h_i$ and $h_j$ is represented as $t_{ij}^{norm}$, indicating that the time is normalized using min-max scaling to lie between zero and one. This is done to make the reward of reaching the final destination (+1) and the penalty of not moving (-1) comparable to the time taken in every step.
\end{enumerate}

For training the agents, DeliverAI relies on Q-learning. Q-learning is an off-policy and model-free RL algorithm that learns the Q function (or the action-value function). This Q-function assigns a Q-value to every state-action pair. Hence, the Q function directly approximates the optimal action policy that the agent should follow to get the maximum reward. In every time step $t$, the Q function of $agent_i$ is updated as:
\[
    Q_{i}(s_{t},a_{t}) \leftarrow Q_{i}(s_{t},a_{t}) + \alpha[r_i(a_t|s_t) 
\]
\begin{equation} \label{eq:qlearning}
    \hspace{80pt} + \gamma \underset{a}{\max}Q(s_{t+1},a) - Q(s_{t},a_{t})]
\end{equation}

The last piece of the training process is deciding how the actions are chosen during learning, i.e. the policy ($\pi)$ that the agents use to explore the environment. This policy is a crucial component as it will determine the solution our agents converge to. We experiment with 2 different policies - Boltzmann Exploration ($\pi_{boltzmann}$) and Epsilon Greedy ($\pi_{epsilon}$). In $\pi_{boltzmann}$, the action is chosen by the Boltzmann probability distribution (often referred to as soft-max probability distribution in literature). In this policy, the higher the Q-value of an action, the higher its probability of being selected as the next action. Hence, the actions that are known to give a higher reward in the future are chosen more frequently based on the temperature parameter $T$. Based on this probability distribution, an action is sampled for each step. For $agent_i$, the probability distribution for choosing $a_t$ given that the current state ($s_t$ ) is given by:
    \[ P(a_t = h_i|s_t) = \frac{e^{\frac{-Q_i(s_t,h_i)}{T}}}{\sum_{a_i \in A}{e^{\frac{-Q_i(s_t,a_i)}{T}}}}  \]
In $\pi_{epsilon}$, the agent chooses a random action with a probability of $\epsilon$, and with a probability of $1-\epsilon$, the agent chooses greedily i.e. the action with the maximum Q-value. For $agent_i$ and a chosen value of $\epsilon$, the action ($a_t$) is chosen by randomly sampling a probability $p$:
    \[
        a_t = \begin{cases}
        \mbox{random } a \in A &  \mbox{if } p < \epsilon\\
        \underset{a}{\mbox{argmax}} \ Q_i(s_{t}, a) & \mbox{if } p \ge \epsilon
      \end{cases}
    \]

\begin{figure*}[h]
    \centering
    \begin{minipage}[t]{\linewidth} % Adjust the width as needed
        \centering
        \includegraphics[width=\linewidth]{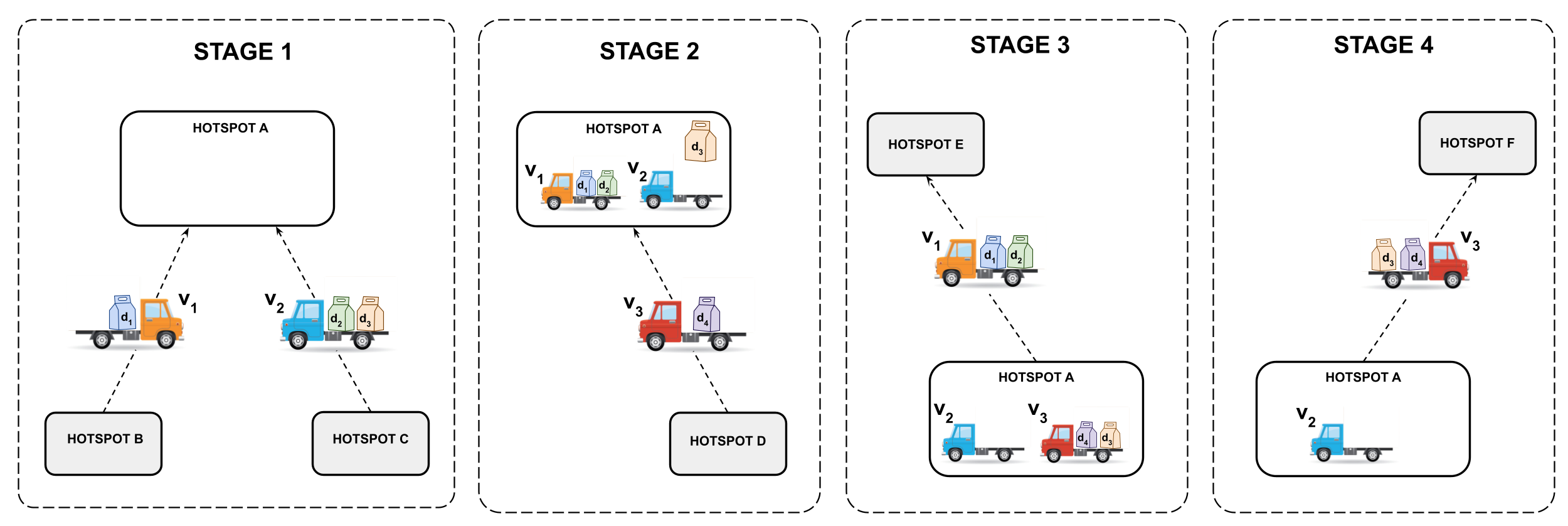}
    \end{minipage}
    \caption{Figure shows the events in a section of the Overlay Network to show how deliveries synchronize for forming a delivery sharing pair. 4 deliveries ($d_1,d_2,d_3,d_4$) traverse through the shown section using 3 vehicles ($v_1,v_2,v_3$). Without sharing, 4 vehicles would be required to traverse the same section.}
    \vspace{2pt}
    \begin{minipage}{0.85\linewidth}
        \footnotesize
        \begin{itemize}
            \item[\textbf{STAGE 1:}] Vehicles $v_1$ and $v_2$ travel to hotspot $A$ carrying deliveries $d_1$ and ($d_2$,$d_3$), respectively.
            \item[\textbf{STAGE 2:}] Deliveries $d_2$ and $d_3$ stop sharing due to the lack of a common hotspot in their $PAS$. Deliveries $d_1$ and $d_2$ form a new sharing pair, while $d_3$ waits for the arrival of $d_4$.
            \item[\textbf{STAGE 3:}] Deliveries ($d_1,d_2$) move to their common hotspot $E$ in vehicle $v_1$. Deliveries $d_3$ and $d_4$ form a sharing pair.
            \item[\textbf{STAGE 4:}] Deliveries ($d_3,d_4$) travel to their common hotspot $F$ in vehicle $v_3$. Vehicle $v_2$ sits idle at hotspot $A$ until it is allocated to another delivery.
        \end{itemize}
    \end{minipage}
    \label{fig:delivery-sharing}
\end{figure*}

\begin{algorithm}
\caption{Q-Learning Algorithm for Training Agents}\label{alg:training}
\small
\begin{algorithmic}[1]
    % \Require {learning rate $\alpha$ $\in$ $(0,1]$, discount rate $\gamma$ $\in$ $(0,1]$ }
    \STATE \text{\textbf{Require:} learning rate $\alpha$ $\in$ $(0,1]$, discount rate $\gamma$ $\in$ $(0,1]$}
    \FOR{\text{each $h_{i}$ $\in$ H}}
        \STATE \text{Initialize $agent_i$ for $h_{i}$} 
        \STATE \text{Initialize $Q_{i}(s, a)$ arbitrarily, for all $s $ $\in$ $S$, $a$ $\in$ $A$} 
        \STATE \text{Initialize policy $\pi$ ($\pi_{epsilon}/ \pi_{boltzmann}$) with respect to $Q_{i}$} 
        \FOR{\text{each episode}}
            \STATE \text{Randomly initialise initial state $s_{0}$ $\neq$ $h_{i}$}
             \STATE \text{Select action $a_{0}$ according to policy $\pi$}
            \FOR{\text{$t = 1,2 .....$ }}
                \STATE \text{Choose $a_{t}$ from $s_{t}$ using $\pi$}
                \STATE \text{Execute action: $s_{t+1} \leftarrow a_t$}
                \STATE \text{$Q_{i}(s_{t},a_{t}) \leftarrow Q_{i}(s_{t},a_{t}) + \alpha[r_i(a_t|s_t)\ +$}
                \STATE \quad \quad \quad \quad \quad \quad \text{$\gamma \ \underset{a}{\max}Q_{i}(s_{t+1},a) - Q_{i}(s_{t},a_{t})]$}
                \IF{$s_{t+1} = h_{i}$}
                    \STATE \text{end episode}
                \ENDIF
            \ENDFOR
        \ENDFOR
    \ENDFOR
\end{algorithmic}
\end{algorithm}

\section{DeliverAI Algorithm}
\label{sec:deliverai}
% \textcolor{purple}{As described in Section X, DeliverAI works in a multi-hop manner. When a delivery D reaches a particular hotspot, our reinforcement learning-based DeliverAI routing algorithm kicks in to determine the best next hop for D. DeliverAI routing takes place in two steps - Agent Interaction and Request Handling. 
% Agent Interaction: During the first step, the agent in charge of delivery D, communicates with all the peers in its transmission range (R) and exchanges their Preferred Action Sets (PAS). A PAS contains a list of the outgoing edges for a delivery ordered by their Q-values. Basically, PAS ranks the best next hops for a delivery (considering at most 10\% of all the outgoing edges) w.r.to the pre-calculated Q-values of the hotspots. When the delivery agent for D receives the PAS of the neighboring peers (delivery agents) it finds out the common next hops between D and with each PAS received. If between delivery D and E, we have multiple common next hops, the respective Q-values for each are added and ranked to choose the best next hotspot to travel to and meet. This prospective meeting of two agents is proposed to a central Request Handler as a REQUEST message which is represented as a quadruple: (D, E, Name of Common Hotspot to meet, Corresponding total Q value).  
% Request Handling: The Request Handler receives all such requests from every potential delivery pair and ranks them based on the Q-sum to allow pairing of the delivering sporting the highest Q-sum.}

As described in Section \ref{sec:delivery-network}, DeliverAI works in a multi-hop manner. When a delivery $d_i$ reaches a particular hotspot, our Reinforcement Learning-based routing algorithm kicks in to determine the best next hop for $d_i$. DeliverAI routing takes place in two steps - \textit{Agent Interaction} and \textit{Request Handling}. 

\subsection{Agent Interaction}
\label{sec:agent-interaction}
During the first step, the agent in charge of delivery $d_i$ ($agent_{d_i^{dest}}$), communicates with all the peers in its transmission range ($R_{agent}$) to exchange their \textit{Preferred Action Sets} ($PAS$). A $PAS$ contains a list of the outgoing edges for a delivery ordered by their Q-values. Basically, $PAS$ ranks the best next hops for a delivery (considering at most 10\% of all the outgoing edges) with respect to the pre-calculated Q-values of the hotspots (Section \ref{sec:training}). Equation \ref{eq:pas} gives a formal notation for the same. Note that the $PAS$ depends not only on the final destination but also on the hotspot ($h$) that $d_i$ communicates from.
\begin{align} \label{eq:pas}
PAS_i(h) =\  & \left\{ a_j \mid  a_j = \underset{a_l}{\mbox{argmax }} Q_{d_i^{dest}}(h,a_l), \right. \nonumber \\
     &\left. a_l \in H \setminus \{a_1, ..., a_{j-1}\},\  j \in \left[1,\frac{|H|}{10}\right] \right\}
\end{align}
When $agent_{d_i^{dest}}$ receives the $PAS$ of the neighboring peers (delivery agents), it finds out the common next hops between $d_i$ and with each $PAS$ received. If between deliveries $d_i$ and $d_j$, we have multiple common next hops, the respective Q-values for each are added and ranked to choose the best next hotspot ($h_{com}$) to travel to and meet (the Q-values are normalized for comparison). This is shown in Equation \ref{eq:common-hotspot}
\begin{align} \label{eq:common-hotspot}
h_{com} =\ & \underset{a}{\mbox{argmax}} \left [Q^{norm}_{d_i^{dest}}(d_i^{curr}(t), a) \ + Q^{norm}_{d_j^{dest}}(d_j^{curr}(t), a) \right] \nonumber \\
& \mbox{where, } a \in PAS_{i} \cap PAS_{j}
\end{align}
\[
    Q_{i}^{norm}(s,a) = \frac{Q_i(s,a) - \underset{a}{\min} (Q_i(s,a))}{\underset{a}{\max}(Q_i(s,a)) - \underset{a}{\min} (Q_i(s,a))} \hspace{10pt} \forall s \in S 
    \]

This prospective meeting of two deliveries is proposed to a central Request Handler as a $Request$ message which is represented as a quadruple: $\left(d_i, d_j, h_{com}, Qsum\right)$, where $Qsum$ is the Q-value sum used in Equation \ref{eq:common-hotspot} for $h_{com}$. The complete algorithm for \textit{Agent Interaction} is shown in Algorithm \ref{alg:agent-interaction}.

% \[
%     a = \underset{a}{\mbox{argmax }}  [Q^{norm}_{d_i^{dest}}(d_i^{next}(t), a) + Q^{norm}_{d_j^{dest}}(d_j^{next}(t), a)]
% \]
% \begin{equation} \label{eq:common-action}
%     a \in P_{i} \cap P_{j} 
% \end{equation}

%     \[ \scriptsize 
%     Q_{i}^{norm}(s,a) = \frac{Q_i^{norm}(s,a) - \underset{a}{\min} (Q_i^{norm}(s,a))}{\underset{a}{\max}(Q_i^{norm}(s,a)) - \underset{a}{\min} (Q_i^{norm}(s,a))} \hspace{10pt} \forall s \in S 
%     \]

\begin{algorithm}
\caption{Algorithm for Agent Interaction}\label{alg:agent-interaction}
\small
\begin{algorithmic}[1]
    \STATE \text{\textbf{Require:} Agent interaction Range: $R_{agent}$,} 
    \STATE \quad \quad \quad \quad \text{Current timestamp: $t$,}
    \STATE \quad \quad \quad \quad \text{\textit{Set} $D_t$: deliveries arriving at a hotspot at time $t$}
    % \STATE \text{\textbf{Ensure:} $d_i$ and $d_j$ are not already sharing}
    \STATE
    \FOR{\text{each $d_i \in D_t$  with $d_i^{share}(t) =$ \textit{NULL}}}
        \STATE \text{Set $PAS_i$ according to Equation \ref{eq:pas}}
        \FOR{\text{each $d_j$ within $R_{agent}$ with $d_j^{share}(t) =$ \textit{NULL}}}
            \STATE \text{Set $PAS_{j}$ according to Equation \ref{eq:pas}}
            \IF{$PAS_{i} \cap PAS_{j} \ne \phi$}
                \STATE \text{Select $h_{com}$ according to Equation \ref{eq:common-hotspot}}
            % \STATE \quad \quad \quad \quad \quad \quad \text{$ + Q_{d_j^{dest}}^{norm}(d_j^{next}(t), a)]$}
                \STATE \text{$Qsum \leftarrow [Q^{norm}_{d_i^{dest}}(d_i^{curr}(t), h_{com})\ +$}  
                \STATE \quad \quad \quad \quad \quad \text{$Q^{norm}_{d_j^{dest}}(d_j^{curr}(t), h_{com})]$}
                \STATE \text{Send $Request$ $(d_i, d_j, h_{com}, Qsum)$}
                \STATE \quad \quad \text{to \textit{Request handler} (Algorithm \ref{alg:handler})}
            \ENDIF
        \ENDFOR
    \ENDFOR
\end{algorithmic}
\end{algorithm}

\subsection{Agent Request Handling}
\label{sec:agent-request-handling}
The \textit{Request Handler} receives all $Requests$ from every potential delivery pair and makes the decision to accept or reject them. Consider that the Request Handler receives 2 requests - $Request_1 = (d_1,d_2,h_i,Qsum_1)$ and $Request_2 = (d_1, d_3, h_j, Qsum_2)$. These requests are conflicting in nature as the delivery $d_1$ cannot travel to $h_i$ (with $d_2$) and $h_j$ (with $d_3$) at the same time. In this case, one request needs to be accepted while the other needs to be rejected. The Request Handler resolves this by accepting the request with a higher $Qsum$ and rejecting the other requests. This is achieved by ranking all the requests based on their $Qsum$ (using a priority queue) and processing them in order as shown in lines [17-26] in Algorithm \ref{alg:handler}. Once a request is approved by the Request Handler, the deliveries are synchronized to meet at the common hotspot and begin sharing.

The Request Handler also assigns the next common hops for deliveries that have already begun sharing (lines [3-15] in Algorithm \ref{alg:handler}) following the procedure described in Equation \ref{eq:common-hotspot}. Notice that once the deliveries have initiated a sharing arrangement, they will persist in sharing as long as their $PAS$s have a common hotspot because forming a new sharing arrangement has an additional  overhead waiting time, which DeliverAI avoids. Lines [28,29] of Algorithm \ref{alg:handler} also shows how the next hop for the deliveries that cannot share, is determined. These deliveries make their next hop individually. A new CDV is allocated at lines [9,24,29] of Algorithm \ref{alg:handler} from the fleet of CDVs available at the respective hotspots.

\begin{algorithm}
\caption{Algorithm for the \textit{Request Handler}}\label{alg:handler}
\small
\begin{algorithmic}[1]
    \STATE \text{\textbf{Require:} Set $D_t$: deliveries arriving at a hotspot at time $t$}
    \STATE
    \FOR{\text{each $d_i$ $\in D_t$ with $d_i^{share}(t) \ne$ \textit{NULL}}}
        \IF{$d_i^{curr}(t)$ =  $d_{i}^{dest}$ or $d_j^{curr}(t)$ = $d_{j}^{dest}$}
            \STATE \text{$d_i^{share}(t^+) \leftarrow $ \textit{NULL}}
            \STATE \text{$d_j^{share}(t^+) \leftarrow $ \textit{NULL}}
        \ELSE
         \STATE \text{Set $PAS_i$, $PAS_j$ according to Equation \ref{eq:pas}}
            % \STATE \text{$h_{com} \leftarrow \underset{a}{\mbox{argmax}}$ $[Q_{d_i^{dest}}^{norm}(d_i^{next}(t), a)$}
            % \STATE \quad \quad \quad \quad \quad \quad \text{$ + Q_{d_j^{dest}}^{norm}(d_j^{next}(t), a)]$}
            \IF{$PAS_{i} \cap PAS_{j} \ne \phi$}
                \STATE \text{Select $h_{com}$ according to Equation \ref{eq:common-hotspot}}
                \STATE \text{$d_i^{next}(t^+) \leftarrow h_{com}$} 
                \STATE \text{$d_i^{next}(t^+) \leftarrow h_{com}$}
            \ELSE
                 \STATE \text{$d_i^{share}(t^+) \leftarrow $ \textit{NULL}}
                 \STATE \text{$d_j^{share}(t^+) \leftarrow $ \textit{NULL}}
            \ENDIF
        \ENDIF  
    \ENDFOR
    \STATE
    \STATE \text{Initialise max-priority queue $Pairs$}
    \STATE \text{Collect all $Requests$ from Algorithm \ref{alg:agent-interaction} into $Pairs$}
    \STATE
    \WHILE{$Pairs$ is not empty}
        \STATE \text{Extract a $Request$ $(d_i, d_j, h_{com}, Qsum)$ from $Pairs$}
        \IF{ $d_i^{share}(t) \ne$ \textit{NULL} or  $d_j^{share}(t) \ne$ \textit{NULL}}
            \STATE \text{Discard the $Request$}
        \ELSE
            \STATE \text{$d_i^{share}(t^+) \leftarrow d_j $, $d_j^{share}(t^+) \leftarrow d_i $}
            \STATE \text{$d_i^{next}(t^+) \leftarrow h_{com}$, $d_i^{next}(t^+) \leftarrow h_{com}$}
        \ENDIF
    \ENDWHILE
    \STATE
    \FOR{\text{each $d_i$ $\in D_t$}with $d_i^{share}(t) =$ \textit{NULL}}
        \STATE \text{$d_i^{next}(t^+) \leftarrow \underset{a}{\mbox{argmax}} \ Q_{d_{i}^{dest}}(d_i^{curr}(t), a)$}
    \ENDFOR
\end{algorithmic}
\end{algorithm}

\begin{table}[]
\caption{Number of Producer And Consumer Locations In Chicago (OSMStreetMaps)}
\label{tab:OSM-data}
\setlength\tabcolsep{3.5pt}
\centering
\begin{tabular}{|c|ccccccccc|ccccccc|}
  \hline &
  \multicolumn{9}{c|}{CONSUMERS} &
  \multicolumn{7}{c|}{PRODUCERS} \\ \hline
  \multicolumn{1}{|c|}{\begin{sideways}\textbf{Locations \ }\end{sideways}} &
  \multicolumn{1}{c|}{\begin{sideways}4185\end{sideways}} &
  \multicolumn{1}{c|}{\begin{sideways}2364\end{sideways}} &
  \multicolumn{1}{c|}{\begin{sideways}449\end{sideways}} &
  \multicolumn{1}{c|}{\begin{sideways}280\end{sideways}} &
  \multicolumn{1}{c|}{\begin{sideways}105\end{sideways}} &
  \multicolumn{1}{c|}{\begin{sideways}435\end{sideways}} &
  \multicolumn{1}{c|}{\begin{sideways}200\end{sideways}} &
  \multicolumn{1}{c|}{\begin{sideways}59\end{sideways}} &
  \textbf{\begin{sideways}8077\end{sideways}} &
  \multicolumn{1}{c|}{\begin{sideways}1578\end{sideways}} &
  \multicolumn{1}{c|}{\begin{sideways}507\end{sideways}} &
  \multicolumn{1}{c|}{\begin{sideways}518\end{sideways}} &
  \multicolumn{1}{c|}{\begin{sideways}145\end{sideways}} &
  \multicolumn{1}{c|}{\begin{sideways}87\end{sideways}} &
  \multicolumn{1}{c|}{\begin{sideways}52\end{sideways}} &
  \textbf{\begin{sideways}2887\end{sideways}} \\ \hline
  \multicolumn{1}{|c|}{\begin{sideways}\textbf{Type}\end{sideways}} &
  \multicolumn{1}{c|}{\begin{sideways}residential\end{sideways}} &
  \multicolumn{1}{c|}{\begin{sideways}apartments\end{sideways}} &
  \multicolumn{1}{c|}{\begin{sideways}house\end{sideways}} &
  \multicolumn{1}{c|}{\begin{sideways}detached\end{sideways}} &
  \multicolumn{1}{c|}{\begin{sideways}hotel\end{sideways}} &
  \multicolumn{1}{c|}{\begin{sideways}office\end{sideways}} &
  \multicolumn{1}{c|}{\begin{sideways}university\end{sideways}} &
  \multicolumn{1}{c|}{\begin{sideways}hospital\end{sideways}} &
  \textbf{\begin{sideways}Total\end{sideways}} &
  \multicolumn{1}{c|}{\begin{sideways}restaurant\end{sideways}} &
  \multicolumn{1}{c|}{\begin{sideways}cafe\end{sideways}} &
  \multicolumn{1}{c|}{\begin{sideways}fast food\end{sideways}} &
  \multicolumn{1}{c|}{\begin{sideways}bakery\end{sideways}} &
  \multicolumn{1}{c|}{\begin{sideways}{supermarket \ }\end{sideways}} &
  \multicolumn{1}{c|}{\begin{sideways}food court\end{sideways}} &
  \textbf{\begin{sideways}Total\end{sideways}} \\ \hline
\end{tabular}
\end{table}

\section{Experimental Setup And Performance Metrics}

\subsection{Data Source}
Due to the dearth of readily available food delivery data sets, we also performed extensive work in generating real-world food delivery data sets for the city of Chicago. We first identified different types of consumers and producers in Chicago using OpenStreetMap (OSM) \cite{OpenStreetMap} - an open-source and freely licensed geographic database maintained and updated regularly by the community. The different categories of consumers and producers, along with their frequencies, are listed in Table \ref{tab:OSM-data}.  To fetch the geographic coordinates (latitude and longitude) of consumers and producers, we used Overpass API \cite{OverpassAPI} (a popular read-only API service native to OSM) that allowed us to query the required data. The boundaries of census tracts in Chicago were sourced from the Chicago data portal \cite{CensusData}. We used the popular GeoPandas library in Python to manage all the data. A dedicated hotspot was placed in every census tract following the procedure explained in Section \ref{sec:delivery-network}. Some of the census tracts, along with the consumer and producer locations, are shown for reference in Figure \ref{fig:hotspot-location}. Further, we acquired the distance and time data for the edges of our Overlay Network using Grasshopper Directions API \cite{GraphHopper} (an open-source routing library). We collected the distance and time data for 30th March 2023, 6:00 PM to 7:00 PM. We chose this date and time because it represents the traffic conditions of a typical weekday during peak demand hours for dinner. All this data is available at XXX.

\begin{figure}
    \centering    
    \includegraphics[width=\linewidth]{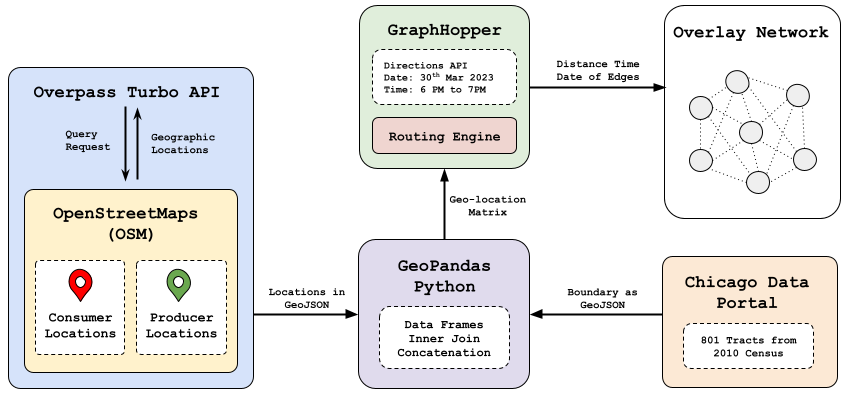}
    \caption{Data Collection Procedure for Simulator Testing. The figure shows how the data is assembled from various sources, organized, and filtered to construct the simulation environment.}
    \label{fig:hotspot-location}
\end{figure}

\begin{figure}
    \centering    
    \includegraphics[width=0.9\linewidth]{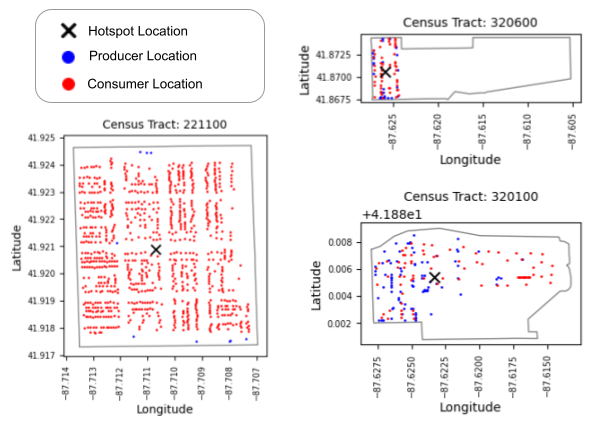}
    \caption{Hotspot placement for a few Census Tracts of Chicago. The hotspot is placed at the centroid of all consumers as explained in Section \ref{sec:delivery-network}.}
    \label{fig:hotspot-location}
\end{figure}

\subsection{Simulation Setup}
\begin{figure}
    \begin{minipage}[t]{0.49\linewidth} % Adjust the width as needed
        \centering
        \vspace{10pt}
        \includegraphics[width=\linewidth]{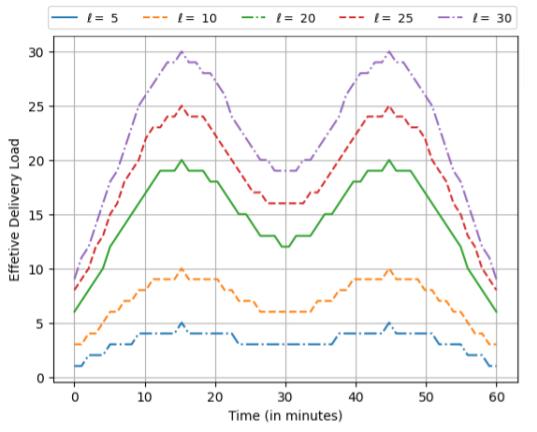}
        % \caption*{(a) Variation of delivery load with respect to time.}
    \end{minipage}
    \begin{minipage}[t]{0.49\linewidth} % Adjust the width as needed
        \centering
        \vspace{10pt}
        \includegraphics[width=\linewidth]{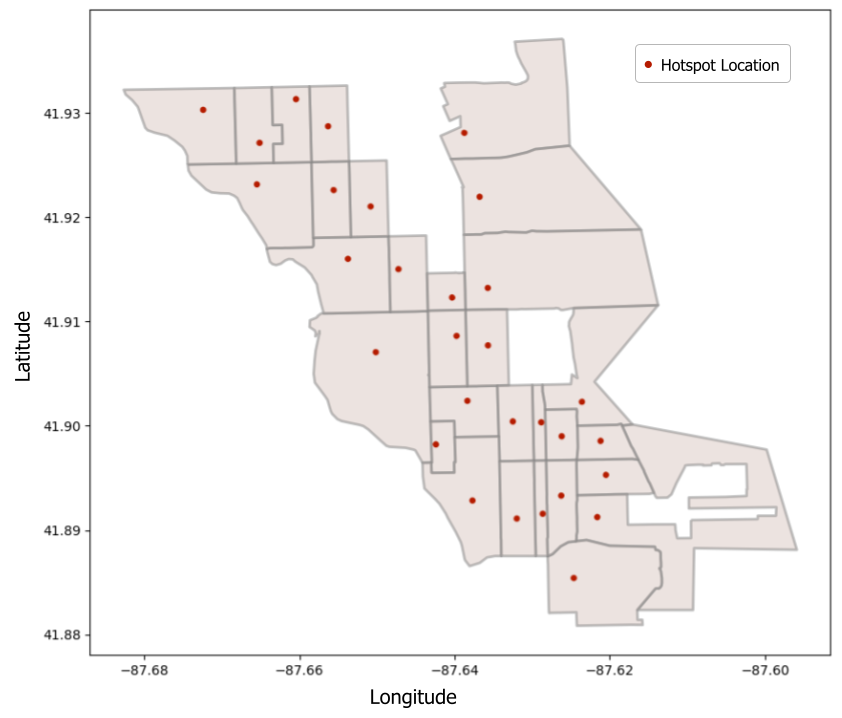}
        % \caption*{(b) $VEH_{tot}$}
    \end{minipage}
    \caption{(a) Variation of delivery load with respect to time. Horizontal lines represent the constant delivery load of Uniform data sets. The curves represent the variation of delivery load in Gaussian data sets. In the Gaussian data sets, observe that the peak delivery demands occur at $t$ = 15 min and 45 min. (b) Figure shows the 30 census tracts in the city of Chicago selected for the simulations. The 992 consumer locations and 356 producer locations lie in the shaded region}
    \label{fig:temp}
\end{figure}

For running the simulations, we chose a region (Figure \ref{fig:temp}) of the Chicago metropolitan area containing 992 consumer and 356 producer locations spanning over 30 census tracts. Due to the want of standard baseline data sets, we construct 2 categories of data sets - \textbf{Uniform} and \textbf{Gaussian}. In the Uniform data sets, we keep the delivery load constant throughout the simulation (Equation \ref{eq:uniform-load}), whereas, in Gaussian data sets, we vary the delivery load according to the Bimodal Gaussian distribution (Equation \ref{eq:gaussian-load}). The parameter $l_o$ is varied to take the values from  {5,10,20,25,30}. The variation of the delivery load as a function of time is also represented in Figure \ref{fig:temp}. The number of deliveries for different values of the parameter $l_o$ is depicted in Table \ref{tab:delivery-frequency}. These data sets simulate real-life conditions and test DeliverAI under varying workloads (number of deliveries). In all the data sets, the required number of deliveries is sampled randomly by selecting a consumer and producer pair randomly. Additionally, since the data sets are sampled randomly, we have constructed 5 different data sets under each category listed in Table \ref{tab:delivery-frequency} to ensure that the results are reproducible.

\begin{equation}
\label{eq:uniform-load}
    l_{uniform}(t) = l_0 \hspace{10pt} \forall \hspace{2pt} t
\end{equation}
\begin{equation}
\label{eq:gaussian-load}
  \begin{aligned}
    l_{gaussian}(t) = \left\lfloor\frac{y(t)}{\underset{t}{\max}(y(t))}\ast \ l_0\right\rfloor \hspace{10pt} \mbox{where,}\\
    y(t) = \frac{1}{\sigma\sqrt{2\pi}}e^{-\frac{1}{2}\left(\frac{t-15}{\sigma}\right)^2} + \frac{1}{\sigma\sqrt{2\pi}}e^{-\frac{1}{2}\left(\frac{t-45}{\sigma}\right)^2} \hspace{5pt} (\sigma = 8) \\
  \end{aligned}
\end{equation}

\renewcommand{\arraystretch}{1.1}
\begin{table}[t]
\centering
\caption{Simulation parameters}
\setlength\tabcolsep{3.5pt}
\centering
\begin{tabular}{|l|l|}
\hline
Parameter                                                           & Value                        \\
\hline
\hline
City Traffic Data used for Date                                     & March 30, 2023 \\ 
\hline
City Traffic Data used for Time                                     & 6 PM to 7 PM  \\ 
\hline
Number of Consumers                                                 & 992                          \\ 
\hline
Number of Producers                                                 & 356                          \\ 
\hline
Simulation Duration ($T_{sim}$)                                     & 60 minutes                   \\ 
\hline
Number of hotspots ($|H|$)                                          & 30                           \\ 
\hline
Number of Agents ($|A|$)                                            & 30                           \\ 
\hline
Size of Preferred Action Set ($|PAS|$)                              & 3                            \\ 
\hline
Agent Interaction Range ($R_{agent}$)                               & 1 Km                  \\ 
\hline
Ideal Delivery Time Window ($\tau$)                                 & 15 minutes                   \\
\hline
$\alpha$ (Equation \ref{eq:qlearning})& 0.8                          \\ 
\hline
$\gamma$ (Equation \ref{eq:qlearning})& 0.99                         \\ 
\hline
$T$ ($\pi_{boltzmann}$)                                             & 10                           \\ 
\hline
$\epsilon$ ($\pi_{epsilon}$)                                        & 0.8                          \\ 
\hline
\end{tabular}
\end{table}

\renewcommand{\arraystretch}{1.1}
\begin{table}[t]
\setlength\tabcolsep{3.5pt}
\caption{Number of deliveries in Uniform and \\ Gaussian Data Sets}
\label{tab:delivery-frequency}
\centering
\begin{tabular}{|c|c|c|c|c|c|} 
\hline
Category & $l_o$=5 & $l_o$=10 & $l_o$=20 & $l_o$=25 & $l_o$=30  \\ 
\hline
Uniform  & 300 & 600  & 1200 & 1500 & 1800  \\ 
\hline
Gaussian & 202 & 432  & 896  & 1130 & 1358  \\
\hline
\end{tabular}
\end{table}

\subsection{Q-learning Implementation }
\label{sec:implementation}
\begin{figure}
    \centering    \includegraphics[width=\linewidth]{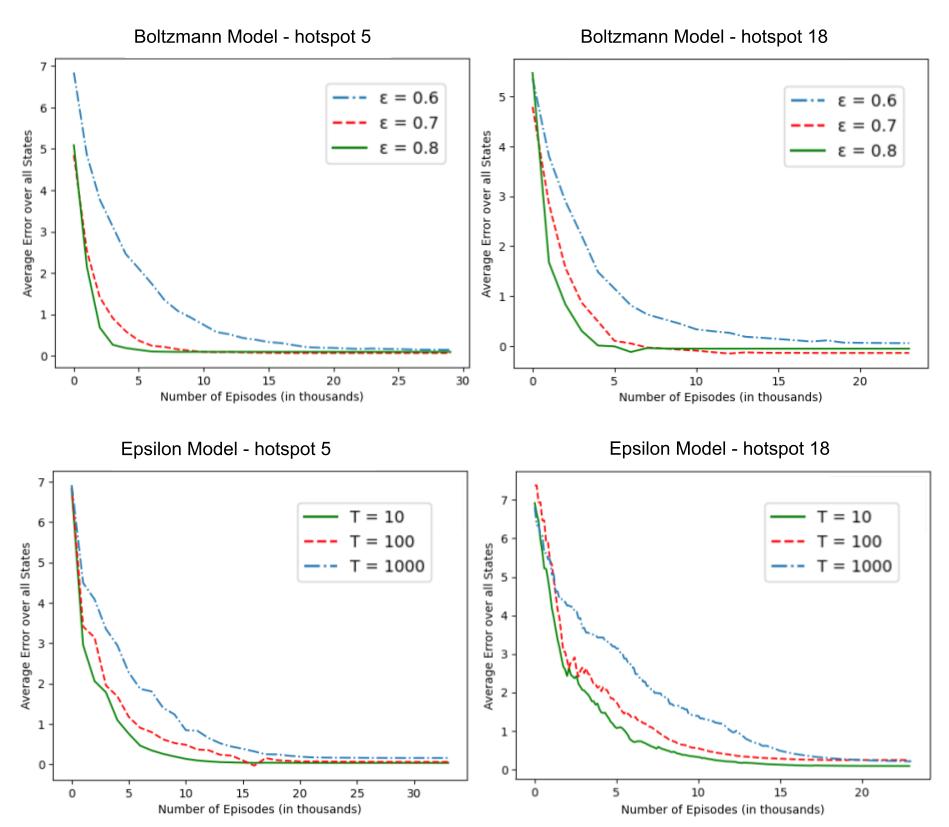}
    \caption{Average Error during Training for Boltzmann Model and Epsilon Model for different values of $\epsilon$ and $T$. $agent_5$ and $agent_{18}$ are chosen for representation. Other agents also follow similar trends.}
    \label{fig:error-plots}
\end{figure}

As explained in Section \ref{sec:training}, we have tested DeliverAI's performance with 2 different policies for training the agents - $\pi_{epsilon}$ and $\pi_{boltzmann}$. For further text, we will refer to them as \textbf{DeliverAI-I} and \textbf{DeliverAI-II}, respectively. Figure \ref{fig:error-plots} shows the training error (averaged over all states) for DeliverAI-I and II during the training procedure described in Section \ref{sec:training}. In DeliverAI-I, lower temperature ($T$) speeds up convergence by favoring high-reward actions. In contrast, DeliverAI-II converges faster with a higher epsilon ($\epsilon$) value which introduces more randomness for exploration rather than exclusively focusing on the best-known sub-optimal action. Hence both models exhibit faster convergence when prioritizing exploration over exploitation. However, DeliverAI-I strikes a better balance and converges more quickly than DeliverAI-II. This is because DeliverAI-I uses $\pi_{boltzmann}$ to randomize action selection, giving more weight to superior actions. In contrast, DeliverAI-II's $\pi_{epsilon}$ selects actions randomly with equal weights for exploration and only exploits the best-known action. We will further do an in-depth comparative analysis of the performance of DeliverAI-I and II with respect to our path-sharing objectives.

\subsection{Baseline Construction}
To compare the results of DeliverAI, we construct 2 baseline solutions summarized below:

\subsubsection{\textbf{Baseline 1}}
This model uses a point-to-point delivery system wherein the delivery is picked up from the producer and dropped off directly to the consumer using a dedicated delivery vehicle (single hop). For this, we route the nearest vehicle to the producer's location within the census tract (to avoid considering far-off vehicles). In case a vehicle is not available within the same census tract, we add a vehicle to the system and make it available for use at the producer's location. The vehicle waits at the consumer location after delivering until assigned to another pickup. At the end of the simulation, we calculate the total number of vehicles as the sum of the maximum number of vehicles needed in every census tract. The delivery time and distance encompasses both the vehicle's travel to the producer and the subsequent journey to the consumer.

\subsubsection{\textbf{Baseline 2}}
This model is analogous to DeliverAI, except that it does not allow delivery sharing. Each delivery in this model travels individually and hence there is no communication between the agents. This baseline helps in analyzing the performance change by introducing the path-sharing mechanism adopted by DeliverAI. All performance metrics are calculated as done for DeliverAI in \ref{sec:performance-metrics}. It is important to note that in baseline 2, all deliveries will take 3 hops - one from the producer to its hotspot, one from the producer's hotspot to the consumer's hotspot, and the last from the consumer's hotspot to the consumer.

\subsection{Performance Metrics}
\label{sec:performance-metrics}
We propose the following metrics to test the performance of DeliverAI. These metrics have been carefully designed to benefit both the consumers, producers and the delivery company.

\begin{enumerate}
    \item \textbf{Total Distance Covered} ($DIST_{tot}$): Measures the sum of the total distance (in km) traveled by each delivery vehicle (CDV and PDV) for the entire duration of the simulation. Lower $DIST_{tot}$ implies lower delivery costs for the company by cutting down on fuel consumption, which ultimately reduces prices for consumers.

    \[
        DIST_{tot} = \sum_{i} v_{i}^{dist}
    \]

    \item \textbf{Average Delivery Time} ($TIME_{avg}$): Average time (in seconds) taken to fulfill a delivery. $TIME_{avg}$  includes 2 components - the time spent traveling in a delivery vehicle  and the time spent at a hotspot waiting for another delivery. Lower $TIME_{avg}$ is necessary to maintain higher customer satisfaction.
    \[
        TIME_{avg} = \sum_{i} \frac{(d_{i}^{end} - d_{i}^{start})}{|D|}
    \]

    \item \textbf{Total Vehicle Requirement} ($VEH_{tot}$): The total number of delivery vehicles (CDVs and PDVs) required in the city to complete all deliveries. Rather than taking in a fixed size of the delivery vehicle fleet, our model has the ability to decide the optimal number of delivery vehicles required for completing all the deliveries. The task of calculating $VEH_{tot}$ is divided into smaller units - calculating the total number of delivery vehicles required in each census tract, which are summed up for calculating $VEH_{tot}$. The number of vehicles in each census tract is taken to be the maximum active vehicles in it throughout the simulation duration. Lower $VEH_{tot}$ will help the delivery company cut down on delivery costs while ensuring that all deliveries are completed.

    \[
        VEHICLE_{tot} = \sum_{i}{\max_{t}{(h_i^{act}(t)})}
    \]

    \item \textbf{Average Number of Hops} ($HOPS_{avg}$): Measures the average (over all the deliveries) number of hops made per delivery. $HOPS_{avg}$ can be linked inversely to the safety of the delivery. Lower $HOPS_{avg}$ is desirable as it reduces the overhead waiting time required for forming delivery sharing pairs. Additionally, it reduces the chances of food contamination associated with multiple handovers between delivery vehicles. Higher $HOPS_{avg}$ also indicates that path-sharing is more frequent.

\[
    HOP_{avg} = \frac{\sum_{i}{(|d_{i}^{path}|)}}{|D|}
\]

    \item \textbf{Utilization Ratio} ($UR_{t}$): Ratio of the number of active delivery vehicles at time $t$ to $VEH_{tot}$. $UR_{t}$ is a measure of the quality of resource (delivery vehicle) allocation.  Higher $UR_t$ indicates optimal vehicle allocation. It ensures that the deliveries are well-distributed among all the delivery vehicles (rather than having a few delivery vehicles make a larger portion of the deliveries). For analysis, we also use $UR_{avg}$, which is the $UR_{t}$ averaged over the stable phase of the simulation ($t$ = 10 min to 50 min).

\[
    UR_{t} = \frac{\sum_{i}{(h_{i}^{act}(t))}}{VEHICLE_{tot}}
\]

\item \textbf{Success Ratio} ($SUCC_{ratio}$): $SUCC_{ratio}$ is the ratio of the number of deliveries completed within time $\tau$ to the total number of deliveries. In our simulation, we set $\tau$ = 15 minutes (900 seconds). All deliveries completed in under 15 minutes are considered \textit{successful}; others are considered \textit{delayed}. A lower $TIME_{avg}$ leads to a higher $SUCC_{ratio}$.

\begin{align*}
    SUCC_{del} &= \{ d_{i} \ | \ (d_{i}^{end} - d_{i}^{start}) \le \tau \} \\
    SUCC_{ratio} &= \frac{|SUCC_{del}|}{|D|}
\end{align*}
\end{enumerate}

\renewcommand{\arraystretch}{1.2}
\begin{table*}[ht]
\captionsetup{
    justification=justified,
    font={footnotesize,sc},
    labelfont={bf},
    textfont={rm},
    singlelinecheck=false}
\setlength\tabcolsep{2.0pt}
\caption{Comparative Results for Gaussian Delivery Load Data Sets - Comparison of DeliverAI-I (D1) and DeliverAI-II (D2) with Baseline 1 (B1) and Baseline 2 (B2) on Performance Metrics. DeliverAI-I uses $\pi_{boltzmann}$ while DeliverAI-II uses $\pi_{epsilon}$. \\ ($\% \downarrow$ represents percentage decrease and $\% \uparrow$ represents percentage increase w.r.t the baselines. All results are averaged over the 5 data sets under each category, for absolute values, refer to Figure \ref{fig:gaussian})}
\label{tab:gaussian-table}
\centering
\scalebox{0.75}{
\begin{tabular}{|c||cccc||cccc||cccc||cccc||cccc|}
\hline
\begin{tabular}[c]{@{}c@{}}Performance \\ Metric\end{tabular} &
  \multicolumn{4}{c||}{(A) $DIST_{tot} (\% \downarrow)$} &
  \multicolumn{4}{c||}{(B) $VEH_{tot} (\% \downarrow)$} &
  \multicolumn{4}{c||}{(C) $TIME_{avg} (\% \uparrow)$} &
  \multicolumn{4}{c||}{(D) $UR_{avg}$ (\%)} &
  \multicolumn{4}{c|}{(E) $SUCC_{ratio}$ (\%)} \\ \hline
  \multirow{2}{*}{\begin{tabular}[c]{@{}c@{}}Delivery\\ Load ($l_o$)\end{tabular}} &
  \multicolumn{2}{c|}{D1} &
  \multicolumn{2}{c||}{D2} &
  \multicolumn{2}{c|}{D1} &
  \multicolumn{2}{c||}{D2} &
  \multicolumn{2}{c|}{D1} &
  \multicolumn{2}{c||}{D2} &
  \multicolumn{1}{c|}{\multirow{2}{*}{D1}} &
  \multicolumn{1}{c|}{\multirow{2}{*}{D2}} &
  \multicolumn{1}{c|}{\multirow{2}{*}{B1}} &
  \multirow{2}{*}{B2} &
  \multicolumn{1}{c|}{\multirow{2}{*}{D1}} &
  \multicolumn{1}{c|}{\multirow{2}{*}{D2}} &
  \multicolumn{1}{c|}{\multirow{2}{*}{B1}} &
  \multirow{2}{*}{B2} \\ \cline{2-13}
 &
  \multicolumn{1}{c|}{B1} &
  \multicolumn{1}{c|}{B2} &
  \multicolumn{1}{c|}{B1} &
  B2 &
  \multicolumn{1}{c|}{B1} &
  \multicolumn{1}{c|}{B2} &
  \multicolumn{1}{c|}{B1} &
  B2 &
  \multicolumn{1}{c|}{B1} &
  \multicolumn{1}{c|}{B2} &
  \multicolumn{1}{c|}{B1} &
  B2 &
  \multicolumn{1}{c|}{} &
  \multicolumn{1}{c|}{} &
  \multicolumn{1}{c|}{} &
   &
  \multicolumn{1}{c|}{} &
  \multicolumn{1}{c|}{} &
  \multicolumn{1}{c|}{} &
   \\ \hline
5 &
  \multicolumn{1}{c|}{7.21} &
  \multicolumn{1}{c|}{13.60} &
  \multicolumn{1}{c|}{7.12} &
  13.51 &
  \multicolumn{1}{c|}{-17.58} &
  \multicolumn{1}{c|}{0.11} &
  \multicolumn{1}{c|}{-15.84} &
  1.63 &
  \multicolumn{1}{c|}{8.46} &
  \multicolumn{1}{c|}{9.31} &
  \multicolumn{1}{c|}{3.62} &
  4.44 &
  \multicolumn{1}{c|}{25.72} &
  \multicolumn{1}{c|}{24.90} &
  \multicolumn{1}{c|}{18.62} &
  22.40 &
  \multicolumn{1}{c|}{94.06} &
  \multicolumn{1}{c|}{97.13} &
  \multicolumn{1}{c|}{99.20} &
  99.21 \\ \hline
10 &
  \multicolumn{1}{c|}{8.25} &
  \multicolumn{1}{c|}{17.55} &
  \multicolumn{1}{c|}{8.35} &
  17.63 &
  \multicolumn{1}{c|}{-7.62} &
  \multicolumn{1}{c|}{8.27} &
  \multicolumn{1}{c|}{-7.21} &
  8.61 &
  \multicolumn{1}{c|}{12.44} &
  \multicolumn{1}{c|}{8.14} &
  \multicolumn{1}{c|}{10.28} &
  6.06 &
  \multicolumn{1}{c|}{46.62} &
  \multicolumn{1}{c|}{46.05} &
  \multicolumn{1}{c|}{25.45} &
  32.75 &
  \multicolumn{1}{c|}{96.90} &
  \multicolumn{1}{c|}{98.06} &
  \multicolumn{1}{c|}{99.42} &
  99.44 \\ \hline
20 &
  \multicolumn{1}{c|}{9.47} &
  \multicolumn{1}{c|}{18.77} &
  \multicolumn{1}{c|}{9.49} &
  18.79 &
  \multicolumn{1}{c|}{8.19} &
  \multicolumn{1}{c|}{14.76} &
  \multicolumn{1}{c|}{8.47} &
  15.02 &
  \multicolumn{1}{c|}{16.00} &
  \multicolumn{1}{c|}{10.67} &
  \multicolumn{1}{c|}{13.62} &
  8.40 &
  \multicolumn{1}{c|}{62.89} &
  \multicolumn{1}{c|}{62.13} &
  \multicolumn{1}{c|}{25.12} &
  38.26 &
  \multicolumn{1}{c|}{95.74} &
  \multicolumn{1}{c|}{97.23} &
  \multicolumn{1}{c|}{99.29} &
  99.29 \\ \hline
25 &
  \multicolumn{1}{c|}{9.80} &
  \multicolumn{1}{c|}{18.85} &
  \multicolumn{1}{c|}{9.90} &
  18.95 &
  \multicolumn{1}{c|}{8.40} &
  \multicolumn{1}{c|}{14.60} &
  \multicolumn{1}{c|}{9.29} &
  15.44 &
  \multicolumn{1}{c|}{18.27} &
  \multicolumn{1}{c|}{13.19} &
  \multicolumn{1}{c|}{15.53} &
  10.57 &
  \multicolumn{1}{c|}{71.31} &
  \multicolumn{1}{c|}{71.24} &
  \multicolumn{1}{c|}{30.48} &
  40.74 &
  \multicolumn{1}{c|}{94.73} &
  \multicolumn{1}{c|}{96.60} &
  \multicolumn{1}{c|}{99.49} &
  99.49 \\ \hline
30 &
  \multicolumn{1}{c|}{13.20} &
  \multicolumn{1}{c|}{21.10} &
  \multicolumn{1}{c|}{13.23} &
  21.13 &
  \multicolumn{1}{c|}{10.39} &
  \multicolumn{1}{c|}{20.92} &
  \multicolumn{1}{c|}{10.52} &
  21.04 &
  \multicolumn{1}{c|}{17.73} &
  \multicolumn{1}{c|}{13.64} &
  \multicolumn{1}{c|}{15.16} &
  11.16 &
  \multicolumn{1}{c|}{75.68} &
  \multicolumn{1}{c|}{74.64} &
  \multicolumn{1}{c|}{32.80} &
  38.82 &
  \multicolumn{1}{c|}{93.58} &
  \multicolumn{1}{c|}{95.35} &
  \multicolumn{1}{c|}{99.83} &
  99.84 \\ \hline
\end{tabular}}
\end{table*}

\renewcommand{\arraystretch}{1.2}
\begin{table*}[ht]
\captionsetup{
    justification=justified,
    font={footnotesize,sc},
    labelfont={bf},
    textfont={rm},
    singlelinecheck=false}
\setlength\tabcolsep{2.0pt}
\caption{Comparative Results for Uniform Delivery Load Data Sets - Comparison of DeliverAI-I (D1) and DeliverAI-II (D2) with Baseline 1 (B1) and Baseline 2 (B2) on Performance Metrics. DeliverAI-I uses $\pi_{boltzmann}$ while DeliverAI-II uses $\pi_{epsilon}$. \\ ($\% \downarrow$ represents percentage decrease and $\% \uparrow$ represents percentage increase w.r.t the baselines. All results are averaged over the 5 data sets under each category, for absolute values, refer to Figure \ref{fig:uniform})}
\label{tab:uniform-table}
\centering
\scalebox{0.75}{
\begin{tabular}{|c||cccc||cccc||cccc||cccc||cccc|}
\hline
\begin{tabular}[c]{@{}c@{}}Performance \\ Metric\end{tabular} &
  \multicolumn{4}{c||}{(A) $DIST_{tot} (\% \downarrow)$} &
  \multicolumn{4}{c||}{(B) $VEH_{tot} (\% \downarrow)$} &
  \multicolumn{4}{c||}{(C) $TIME_{avg} (\% \uparrow)$} &
  \multicolumn{4}{c||}{(D) $UR_{avg}$ (\%)} &
  \multicolumn{4}{c|}{(E) $SUCC_{ratio}$ (\%)} \\ \hline
  \multirow{2}{*}{\begin{tabular}[c]{@{}c@{}}Delivery\\ Load ($l_o$)\end{tabular}} &
  \multicolumn{2}{c|}{D1} &
  \multicolumn{2}{c||}{D2} &
  \multicolumn{2}{c|}{D1} &
  \multicolumn{2}{c||}{D2} &
  \multicolumn{2}{c|}{D1} &
  \multicolumn{2}{c||}{D2} &
  \multicolumn{1}{c|}{\multirow{2}{*}{D1}} &
  \multicolumn{1}{c|}{\multirow{2}{*}{D2}} &
  \multicolumn{1}{c|}{\multirow{2}{*}{B1}} &
  \multirow{2}{*}{B2} &
  \multicolumn{1}{c|}{\multirow{2}{*}{D1}} &
  \multicolumn{1}{c|}{\multirow{2}{*}{D2}} &
  \multicolumn{1}{c|}{\multirow{2}{*}{B1}} &
  \multirow{2}{*}{B2} \\ \cline{2-13}
 &
  \multicolumn{1}{c|}{B1} &
  \multicolumn{1}{c|}{B2} &
  \multicolumn{1}{c|}{B1} &
  B2 &
  \multicolumn{1}{c|}{B1} &
  \multicolumn{1}{c|}{B2} &
  \multicolumn{1}{c|}{B1} &
  B2 &
  \multicolumn{1}{c|}{B1} &
  \multicolumn{1}{c|}{B2} &
  \multicolumn{1}{c|}{B1} &
  B2 &
  \multicolumn{1}{c|}{} &
  \multicolumn{1}{c|}{} &
  \multicolumn{1}{c|}{} &
   &
  \multicolumn{1}{c|}{} &
  \multicolumn{1}{c|}{} &
  \multicolumn{1}{c|}{} &
   \\ \hline
5 &
  \multicolumn{1}{c|}{7.01} &
  \multicolumn{1}{c|}{17.07} &
  \multicolumn{1}{c|}{6.73} &
  16.81 &
  \multicolumn{1}{c|}{-7.20} &
  \multicolumn{1}{c|}{11.16} &
  \multicolumn{1}{c|}{-7.65} &
  10.80 &
  \multicolumn{1}{c|}{23.19} &
  \multicolumn{1}{c|}{16.40} &
  \multicolumn{1}{c|}{19.83} &
  13.22 &
  \multicolumn{1}{c|}{45.59} &
  \multicolumn{1}{c|}{43.55} &
  \multicolumn{1}{c|}{21.59} &
  27.94 &
  \multicolumn{1}{c|}{91.33} &
  \multicolumn{1}{c|}{94.13} &
  \multicolumn{1}{c|}{99.80} &
  99.84 \\ \hline
10 &
  \multicolumn{1}{c|}{8.38} &
  \multicolumn{1}{c|}{17.90} &
  \multicolumn{1}{c|}{6.98} &
  16.65 &
  \multicolumn{1}{c|}{4.90} &
  \multicolumn{1}{c|}{9.96} &
  \multicolumn{1}{c|}{5.13} &
  10.21 &
  \multicolumn{1}{c|}{17.18} &
  \multicolumn{1}{c|}{11.38} &
  \multicolumn{1}{c|}{14.97} &
  9.28 &
  \multicolumn{1}{c|}{50.12} &
  \multicolumn{1}{c|}{48.78} &
  \multicolumn{1}{c|}{25.45} &
  33.56 &
  \multicolumn{1}{c|}{93.33} &
  \multicolumn{1}{c|}{95.26} &
  \multicolumn{1}{c|}{99.76} &
  99.77 \\ \hline
20 &
  \multicolumn{1}{c|}{10.13} &
  \multicolumn{1}{c|}{19.31} &
  \multicolumn{1}{c|}{10.37} &
  19.52 &
  \multicolumn{1}{c|}{7.53} &
  \multicolumn{1}{c|}{18.91} &
  \multicolumn{1}{c|}{7.13} &
  18.54 &
  \multicolumn{1}{c|}{19.92} &
  \multicolumn{1}{c|}{13.69} &
  \multicolumn{1}{c|}{17.29} &
  11.20 &
  \multicolumn{1}{c|}{69.78} &
  \multicolumn{1}{c|}{68.11} &
  \multicolumn{1}{c|}{32.85} &
  37.75 &
  \multicolumn{1}{c|}{93.55} &
  \multicolumn{1}{c|}{95.43} &
  \multicolumn{1}{c|}{99.20} &
  99.20 \\ \hline
25 &
  \multicolumn{1}{c|}{10.67} &
  \multicolumn{1}{c|}{20.56} &
  \multicolumn{1}{c|}{10.81} &
  20.69 &
  \multicolumn{1}{c|}{8.11} &
  \multicolumn{1}{c|}{20.72} &
  \multicolumn{1}{c|}{8.99} &
  21.49 &
  \multicolumn{1}{c|}{20.98} &
  \multicolumn{1}{c|}{14.33} &
  \multicolumn{1}{c|}{18.63} &
  12.12 &
  \multicolumn{1}{c|}{80.12} &
  \multicolumn{1}{c|}{80.10} &
  \multicolumn{1}{c|}{33.76} &
  40.15 &
  \multicolumn{1}{c|}{94.52} &
  \multicolumn{1}{c|}{96.16} &
  \multicolumn{1}{c|}{99.49} &
  99.51 \\ \hline
30 &
  \multicolumn{1}{c|}{13.46} &
  \multicolumn{1}{c|}{22.43} &
  \multicolumn{1}{c|}{13.17} &
  22.17 &
  \multicolumn{1}{c|}{12.53} &
  \multicolumn{1}{c|}{24.66} &
  \multicolumn{1}{c|}{12.12} &
  24.31 &
  \multicolumn{1}{c|}{20.49} &
  \multicolumn{1}{c|}{14.38} &
  \multicolumn{1}{c|}{19.12} &
  13.07 &
  \multicolumn{1}{c|}{87.26} &
  \multicolumn{1}{c|}{85.32} &
  \multicolumn{1}{c|}{35.40} &
  40.57 &
  \multicolumn{1}{c|}{94.11} &
  \multicolumn{1}{c|}{95.26} &
  \multicolumn{1}{c|}{99.40} &
  99.40 \\ \hline
\end{tabular}}
\end{table*}

\section{Performance Analysis And Results}

Figures \ref{fig:gaussian} and \ref{fig:uniform} along with Tables \ref{tab:gaussian-table} and \ref{tab:uniform-table} provide performance results for DeliverAI-I and II as well as Baseline 1 and Baseline 2 as per the performance metrics described in Section \ref{sec:performance-metrics}.

\subsection{Comparing DeliverAI-I with DeliverAI-II }
Among DeliverAI-I and DeliverAI-II, we conclude that, even though their performances are very similar, the latter performs better across most performance metrics. With respect to $DIST_{tot}$, we see that DeliverAI-I performs better initially. However, DeliverAI-II matches the performance from $l_o = 20$ onward (Tables \ref{tab:gaussian-table}(A) and \ref{tab:uniform-table}(A)). With respect to the $VEHICLES_{tot}$, $TIME_{avg}$, and $SUCC_{ratio}$, DeliverAI-II consistently performs better than DeliverAI-I though the difference between their performances gets smaller as the delivery load parameter increases. The improved performance of DeliverAI-II can be attributed to its epsilon-greedy strategy, which enables it to explore all available paths with equal emphasis. This approach proves advantageous because the path-sharing mechanism of DeliverAI can seamlessly utilize any of these alternative paths. In the $UR_{avg}$, however, we note the better performance of DeliverAI-I. Since DeliverAI-I uses the Boltzmann exploration strategy, it tends to explore a few selected paths more frequently than others, because of which it re-directs more deliveries through these path segments leading to more path-sharing as shown by the higher $HOPS_{avg}$ in DeliverAI-I when compared to DeliverAI-II (Figure \ref{fig:gaussian}(D) and \ref{fig:uniform}(D)). This allows DeliverAI-I to re-utilize vehicles more frequently. Apart from a lower $UR_{avg}$, DeliverAI-II also comes at the cost of a longer training time, as shown in Section \ref{sec:implementation}. Regularly changing traffic conditions might warrant updating agents' policies frequently, wherein DeliverAI-II can encounter slower responses because of longer training time. Thus, there exists a trade-off between training time and (slightly) better performance when choosing between DeliverAI-I and II.

\subsection{Comparing DeliverAI models with Baselines}

\subsubsection{$DIST_{tot}$}
Our results show that the path-sharing mechanism of DeliverAI is able to save delivery costs in both low and high delivery loads, and its use in peak delivery hours (higher $l_o$) is especially beneficial to the delivery company. In Tables \ref{tab:gaussian-table}(A) and \ref{tab:uniform-table}(A), it is evident that as the delivery load increases, there is a significant improvement in the reduction of $DIST_{tot}$ compared to the baseline values. Under peak delivery load conditions ($l_0 = 30$), DeliverAI saves more than 13\% distance w.r.t. Baseline 1 and well above 21\% distance w.r.t. Baseline 2. With a higher delivery load, multiple sharing options are available, and DeliverAI is able to make intelligent decisions to choose only the best sharing pairs and discard the others.

\subsubsection{$TIME_{avg}$}
As expected, DeliverAI increases $TIME_{avg}$ with increasing $l_o$ (Tables \ref{tab:gaussian-table}(C) and \ref{tab:uniform-table}(C)). More path-sharing generally leads to a rise in $TIME_{avg}$ because of waiting (synchronization delays between deliveries) and detours from the original shortest path. However, in our experimentation, we also observe that, in some cases, $TIME_{avg}$ increases slowly or even decreases when the delivery load increases. This is more pronounced when going from $l_0=5$ to $l_0=10$ in the Uniform data sets and $l_0=25$ to $l_0=30$ in both Uniform and Gaussian data sets. This shows that, with more options to choose from, DeliverAI only allows the optimal sharing pairs to be formed, which helps prevent $TIME_{avg}$ from shooting up exponentially. 

\subsubsection{$VEH_{tot}$}
\begin{figure*}[h]
    \centering
    \begin{minipage}[t]{\linewidth} % Adjust the width as needed
        \centering
        \includegraphics[width=1\linewidth]{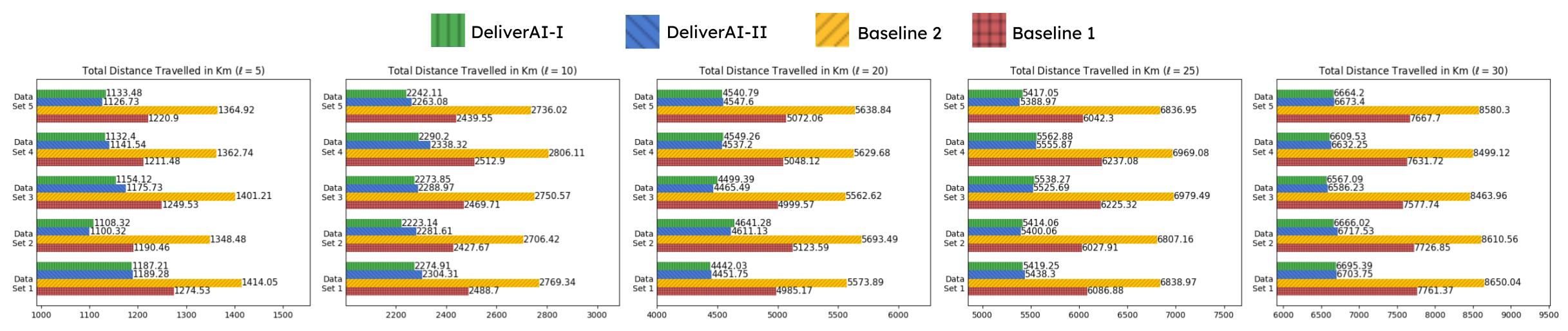}
        \caption*{(A) $DIST_{tot}$ in km}
    \end{minipage}
    \begin{minipage}[t]{\linewidth} % Adjust the width as needed
        \centering
        \vspace{10pt}
        \includegraphics[width=1\linewidth]{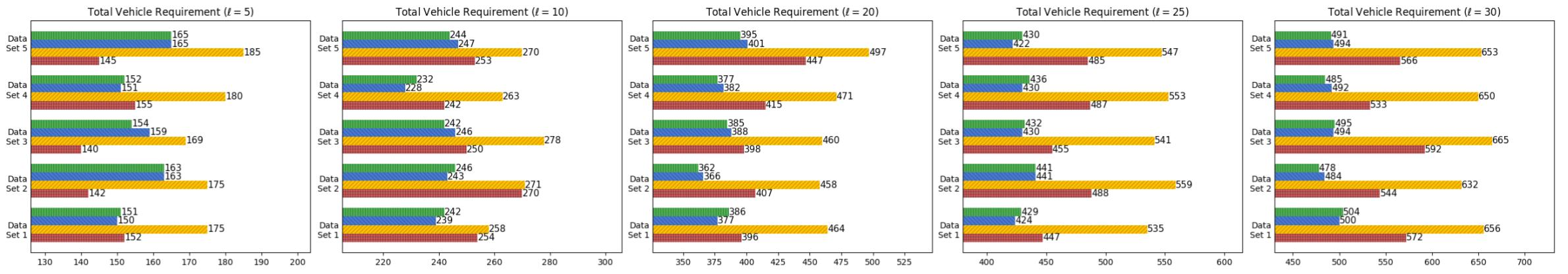}
        \caption*{(B) $VEH_{tot}$}
    \end{minipage}
    \begin{minipage}[t]{\linewidth} % Adjust the width as needed
        \centering
        \vspace{10pt}
        \includegraphics[width=1\linewidth]{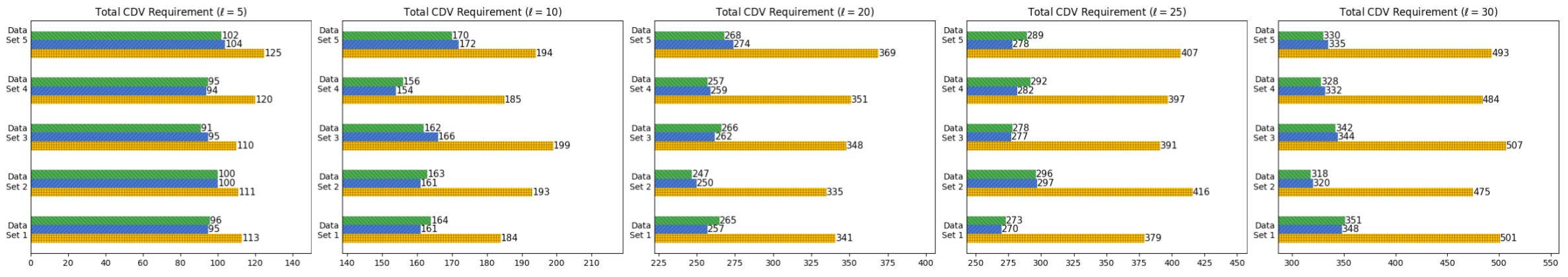}
        \caption*{(C) Total CDV Requirement (Baseline1 has no CDVs hence omitted)}
    \end{minipage}
    \begin{minipage}[t]{\linewidth} % Adjust the width as needed
        \centering
        \vspace{10pt}
        \includegraphics[width=1\linewidth]{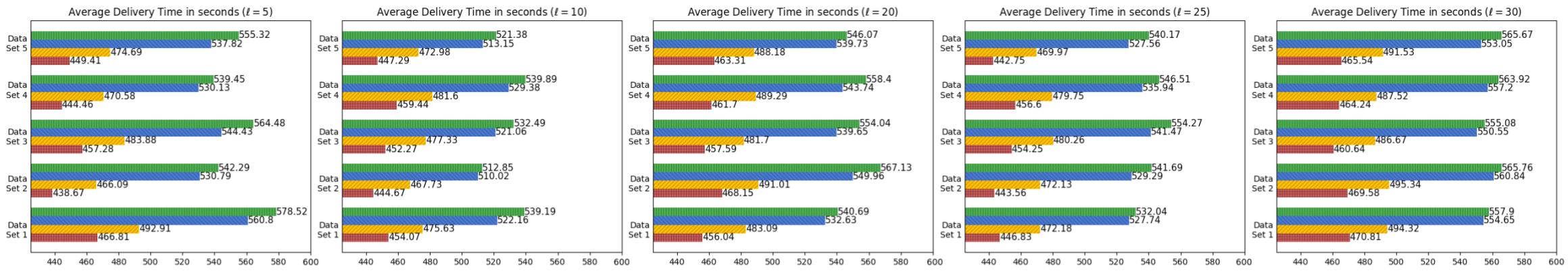}
        \caption*{(D) $TIME_{avg}$ in seconds}
    \end{minipage}
    \begin{minipage}[t]{\linewidth} % Adjust the width as needed
        \centering
        \vspace{10pt}
        \includegraphics[width=1\linewidth]{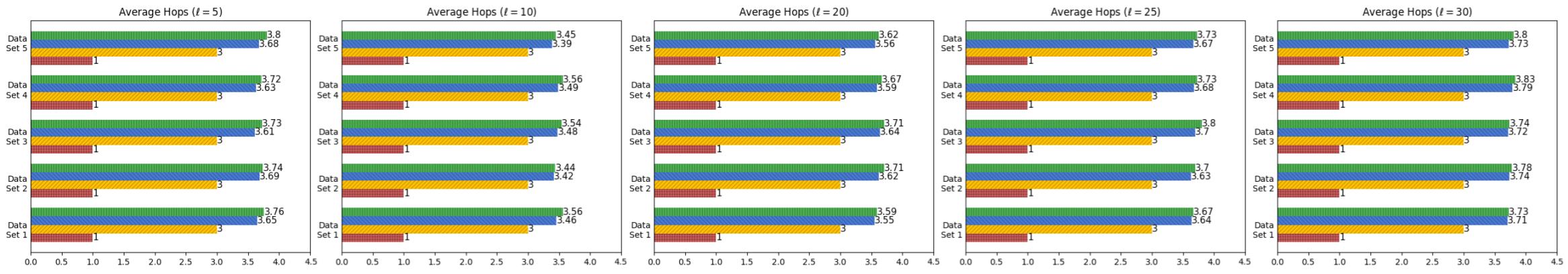}
        \caption*{(E) $HOPS_{avg}$ [Baseline1 and Baseline2 have $HOPS_{avg}=1,3$ (respectively) by definition]}
    \end{minipage}
    \begin{minipage}[t]{\linewidth} % Adjust the width as needed
        \centering
        \vspace{10pt}
        \includegraphics[width=1\linewidth]{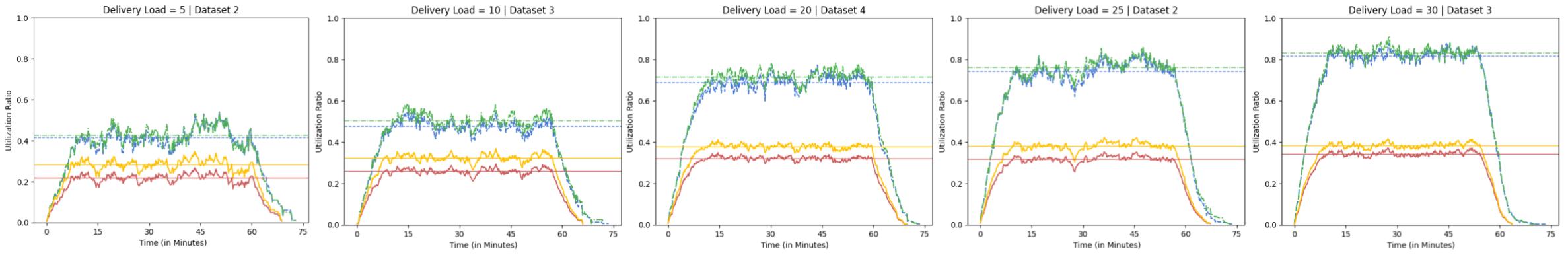}
        \caption*{$UR_{t}$ - Horizontal Lines represent $UR_{avg}$}
        \vspace{10pt}
    \end{minipage}
    \caption{Performance Metrics for Uniform Delivery Load Data Sets varying with Delivery Load ($l_o$) DeliverAI-I uses $\pi_{boltzmann}$ while DeliverAI-II uses $\pi_{epsilon}$}
    \label{fig:uniform}
\end{figure*}

For lower delivery loads ($l_0=5,10$ in Gaussian and $l_0=5$ in Uniform data sets), DeliverAI-I and II require more vehicles than Baseline 1 (as indicated by negative percentages in Tables \ref{tab:gaussian-table}(B) and \ref{tab:uniform-table}(B)). In DeliverAI, we require 2 additional PDVs - one from the producer to a hotspot and another from a hotspot to the consumer - apart from the CDVs in the Overlay Network. This accounts for more vehicles required in DeliverAI for lower delivery loads. However, as $l_o$ increases to 10 in Uniform data sets and 20 in Gaussian data sets, path-sharing becomes more frequent, as demonstrated by the higher $HOPS_{avg}$ in Figure \ref{fig:gaussian}(E) and \ref{fig:uniform}(E). Due to the sharing, CDV requirements drop notably, leading to a lower $VEH_{tot}$ compared to Baseline 1. Our results show that, when compared with Baseline 1, in peak delivery loads ($l_0 = 30$), DeliverAI can reduce $VEH_{tot}$ by over 12\% and 10\% in Uniform and Gaussian data sets, respectively. To compare the effect of path-sharing alone, we look at Baseline 2, which requires significantly higher CDVs (Figure \ref{fig:gaussian}(C) and \ref{fig:uniform}(C)) than DeliverAI-I and II. Due to the lack of path-sharing, the $VEH_{tot}$ (Figure \ref{fig:gaussian}(B) and \ref{fig:uniform}(B)) in Baseline 2 shoots up as $l_o$ increases. On the other hand, the intelligent sharing introduced by DeliverAI in the overlay network results in a remarkable reduction in vehicle requirements, saving more than 20\% and 24\% of $VEHICLES_{tot}$ in Uniform and Gaussian data sets, respectively, compared to Baseline 2.

\subsubsection{$UR_{t}$ and $UR_{avg}$}
\begin{figure*}[h]
    \centering
    \begin{minipage}[t]{\linewidth} % Adjust the width as needed
        \centering
        \includegraphics[width=1\linewidth]{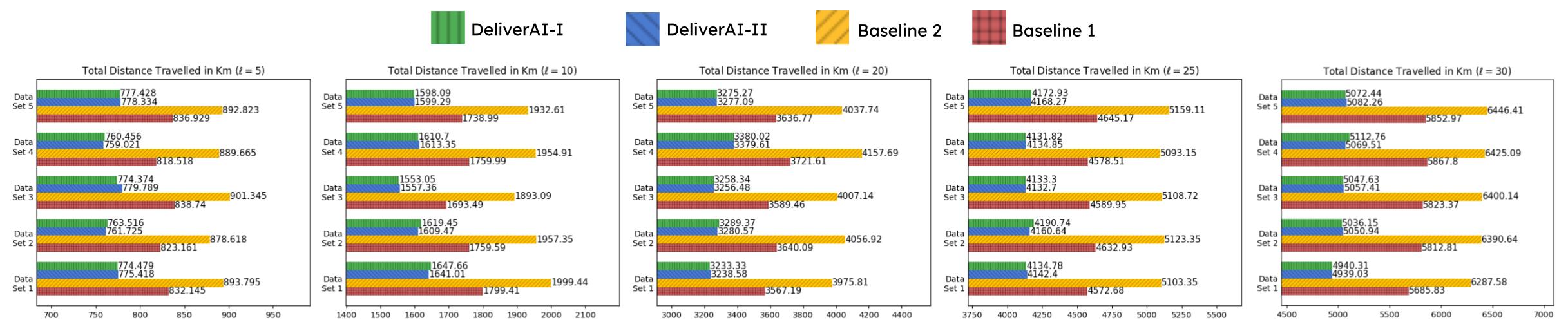}
        \caption*{(A) $DIST_{tot}$ in km}
    \end{minipage}
    \begin{minipage}[t]{\linewidth} % Adjust the width as needed
        \centering
        \vspace{10pt}
        \includegraphics[width=1\linewidth]{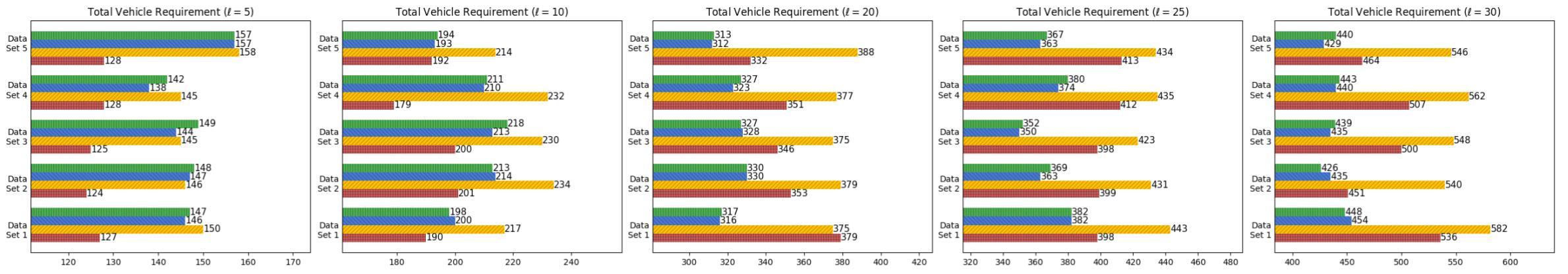}
        \caption*{(B) $VEH_{tot}$}
    \end{minipage}
    \begin{minipage}[t]{\linewidth} % Adjust the width as needed
        \centering
        \vspace{10pt}
        \includegraphics[width=1\linewidth]{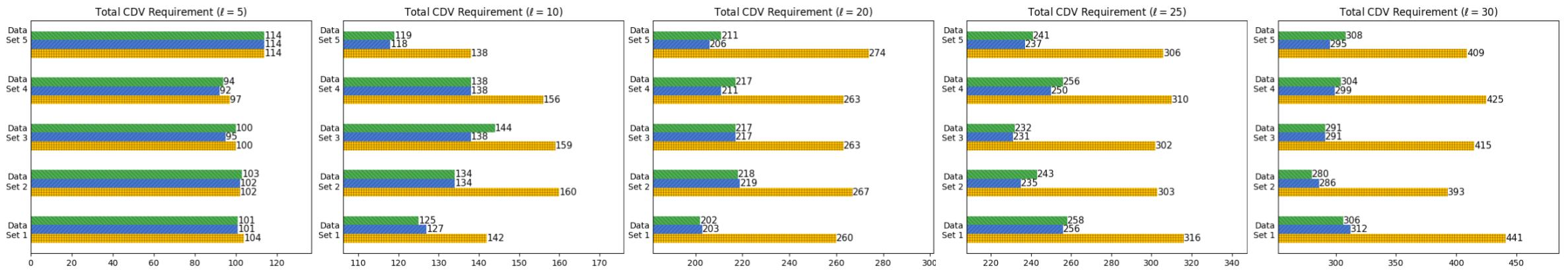}
        \caption*{(C) Total CDV Requirement (Baseline1 has no CDVs hence omitted)}
    \end{minipage}
    \begin{minipage}[t]{\linewidth} % Adjust the width as needed
        \centering
        \vspace{10pt}
        \includegraphics[width=1\linewidth]{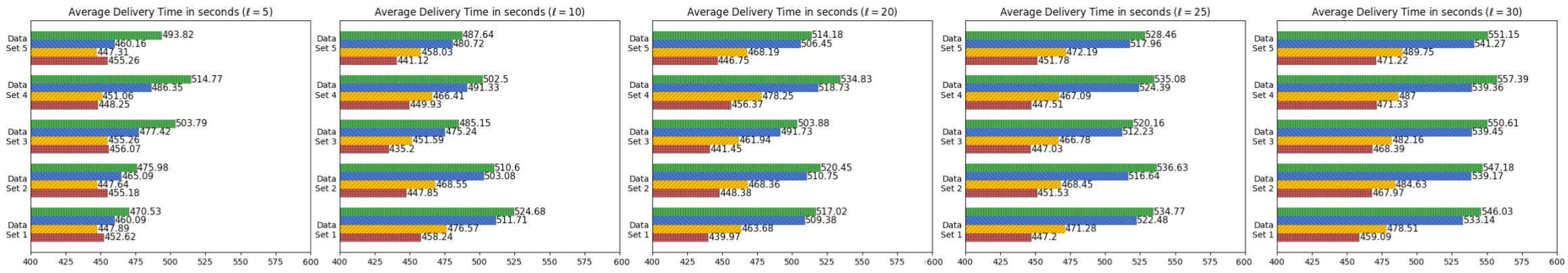}
        \caption*{(D) $TIME_{avg}$ in seconds}
    \end{minipage}
    \begin{minipage}[t]{\linewidth} % Adjust the width as needed
        \centering
        \vspace{10pt}
        \includegraphics[width=1\linewidth]{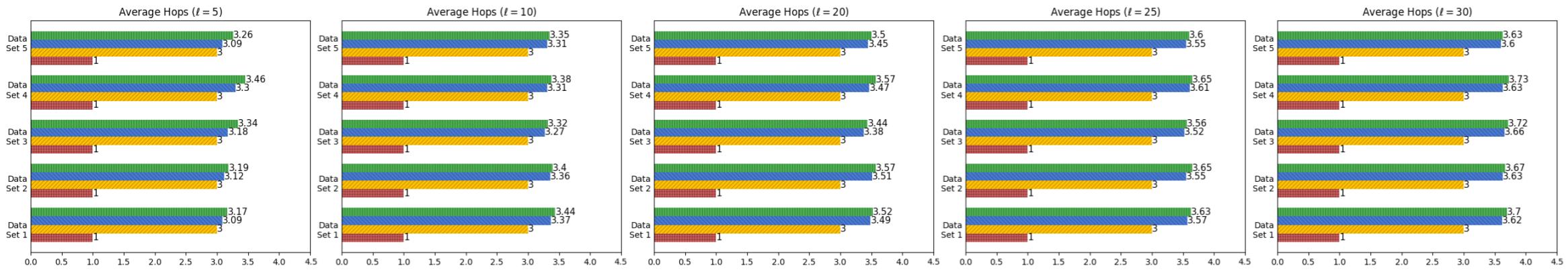}
        \caption*{(E) $HOPS_{avg}$ [Baseline1 and Baseline2 have $HOPS_{avg}=1,3$ (respectively) by definition]}
    \end{minipage}
    \begin{minipage}[t]{\linewidth} % Adjust the width as needed
        \centering
        \vspace{10pt}
        \includegraphics[width=1\linewidth]{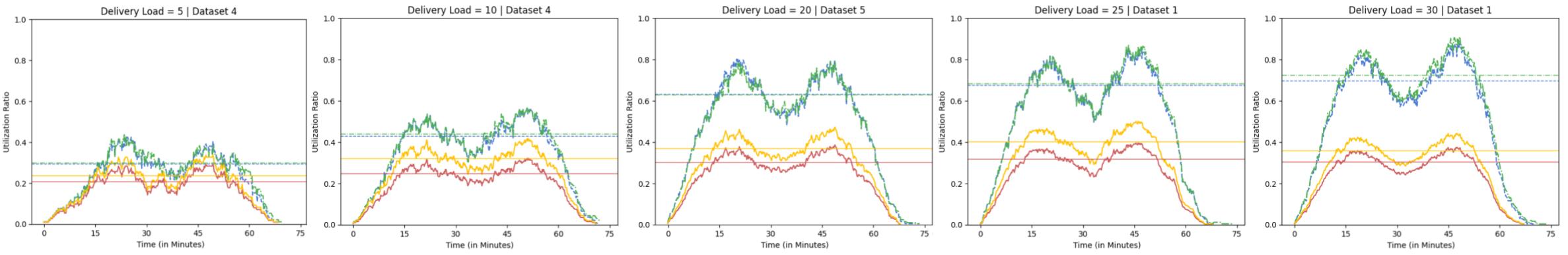}
        \caption*{(F) $UR_{t}$ - Horizontal Lines represent $UR_{avg}$}
        \vspace{10pt}
    \end{minipage}
    \caption{Performance Metrics for Gaussian Delivery Load Data Sets varying with Delivery Load ($l_o$) DeliverAI-I uses $\pi_{boltzmann}$ while DeliverAI-II uses $\pi_{epsilon}$}
    \label{fig:gaussian}
\end{figure*}

As shown in Figures \ref{fig:gaussian}(E) and \ref{fig:uniform}(E), $UR_t$ initially rises as incoming deliveries outnumber completions, leading to an increase in active vehicles. As the incoming and completed deliveries balance out, $UR_t$ stabilizes. Towards the end, deliveries are completed, but new ones stop arriving after the 60-minute mark. Hence, $UR_t$ decreases as the vehicles are freed up. The region that interests us is the part where the $UR_{t}$ is stable. In the Gaussian data sets, this stable phase also depicts 2 peaks (Figure \ref{fig:gaussian}(F)) which coincides with the 2 peaks in the delivery load as shown in Figure \ref{fig:gaussian}. The stable phase depicts the conditions that the delivery company would experience during peak working hours. Tables \ref{tab:gaussian-table}(D) and \ref{tab:uniform-table}(D) show $UR_{avg}$ for this phase. DeliveriAI-I and II offer more than 50\% and 40\% higher $UR_{avg}$ when compared to Baseline 1 in the Uniform and Gaussian data sets, respectively. We also observe that, in comparison with the baselines, $UR_{avg}$ of DeliverAI-I and II increases more rapidly with the increase in $l_o$ (Figure \ref{fig:gaussian}(F) and \ref{fig:uniform}(F)). These trends suggest that our proposed model optimally distributes the additional deliveries across the delivery vehicles, enabling efficient utilization of the existing vehicle network without the need for additional vehicles.  

\subsubsection{$SUCC_{ratio}$}
The $\tau = 15$ minute time window established for the Success Ratio serves as a stringent benchmark, ensuring that both baseline models consistently maintain $SUCC_{ratio}$ over 99\%.  Table \ref{tab:gaussian-table}(E) and \ref{tab:uniform-table}(E) show the $SUCC_{ratio}$ achieved by DeliverAI-I and II and the baselines. As anticipated, DeliverAI exhibits a slightly lower $SUCC_{ratio}$ compared to the baselines across varying delivery loads and data sets. This is because of a $TIME_{avg}$ associated with its advanced path-sharing process. However, it's noteworthy that DeliverAI-II consistently upholds a commendable $SUCC_{ratio}$ approximating 95\% in Uniform and 97\% in Gaussian data sets. Since the deliveries are randomly sampled, there is no direct correlation between the $SUCC_{ratio}$ and the delivery load parameter ($l_o$). But generally, it is observed that the difference between the $SUCC_{ratio}$ of the baselines and DeliverAI gets smaller with larger delivery loads which can be explained similarly as done for the decrease in $TIME_{avg}$.

\section{Conclusion and Future Work}
In this paper, we presented DeliverAI - a Reinforcement Learning-based, distributed path-sharing algorithm that provides dynamic routing for food deliveries. Our main objective behind path-sharing is to reduce the cumulative distance traveled by all deliveries and reduce the fleet size (vehicle requirement). For this, we trained specialized agents to handle the routing of deliveries (hop-by-hop) and developed the Overlay Network to enable Multi-Agent communication. Robust experimentation of DeliverAI on our real data simulator for the city of Chicago has shown the superior performance of DeliverAI over the current models adopted by delivery companies. DeliverAI achieves a significant reduction in the distance traveled by delivery vehicles, optimizes vehicle requirements, and enhances fleet utilization, all while ensuring the timely completion of deliveries by maintaining a high Success Ratio. Currently, our experiments with DeliverAI limit the vehicle capacity to 2 deliveries. Our success with this model has motivated us to incorporate larger vehicle capacities in DeliverAI. We also seek to investigate alternative strategies like Deep Q-Networks (DQN), Double DQN (DDQN), and other multi-agent algorithms to train DeliverAI agents to extend DeliverAI to larger Overlay Networks. Additionally, we aim to quantify our results further by introducing a reward model to estimate the cost-benefit of DeliverAI. 
\newpage

\clearpage
\bibliography{main}

\begin{thebibliography}{10}

\bibitem{GraphHopper}
{The GraphHopper Directions API Route Planning}.

\bibitem{crowddeliver}
Chao Chen, Daqing Zhang, Xiaojuan Ma, Bin Guo, Leye Wang, Yasha Wang, and Edwin Sha.
\newblock Crowddeliver: Planning city-wide package delivery paths leveraging the crowd of taxis.
\newblock {\em IEEE Transactions on Intelligent Transportation Systems}, 18(6):1478--1496, 2016.

\bibitem{deepfreight}
Jiayu Chen, Abhishek~K Umrawal, Tian Lan, and Vaneet Aggarwal.
\newblock Deepfreight: A model-free deep-reinforcement-learning-based algorithm for multi-transfer freight delivery.
\newblock In {\em Proceedings of the International Conference on Automated Planning and Scheduling}, volume~31, pages 510--518, 2021.

\bibitem{mdmpmp}
Wenyi Chen, Martijn Mes, and Marco Schutten.
\newblock Multi-hop driver-parcel matching problem with time windows.
\newblock {\em Flexible services and manufacturing journal}, 30:517--553, 2018.

\bibitem{pptaxi}
Yueyue Chen, Deke Guo, Ming Xu, Guoming Tang, Tongqing Zhou, and Bangbang Ren.
\newblock Pptaxi: Non-stop package delivery via multi-hop ridesharing.
\newblock {\em IEEE Transactions on Mobile Computing}, 19(11):2684--2698, 2019.

\bibitem{CensusData}
{Chicago Data Portal}.
\newblock {Chicago 2010 Census Data}.

\bibitem{dijkstra2022note}
Edsger~W Dijkstra.
\newblock A note on two problems in connexion with graphs.
\newblock In {\em Edsger Wybe Dijkstra: His Life, Work, and Legacy}, pages 287--290. 2022.

\bibitem{dsouza2021online}
Durant Dsouza and Dipasha Sharma.
\newblock Online food delivery portals during covid-19 times: an analysis of changing consumer behavior and expectations.
\newblock {\em International Journal of Innovation Science}, 13(2):218--232, 2021.

\bibitem{CustomerTrends}
Martin Joerss, Florian Neuhaus, and J{\"u}rgen Schr{\"o}der.
\newblock How customer demands are reshaping last-mile delivery.
\newblock {\em The McKinsey Quarterly}, 17:1--5, 2016.

\bibitem{fleetmanagement}
Kaixiang Lin, Renyu Zhao, Zhe Xu, and Jiayu Zhou.
\newblock Efficient large-scale fleet management via multi-agent deep reinforcement learning.
\newblock In {\em Proceedings of the 24th ACM SIGKDD international conference on knowledge discovery \& data mining}, pages 1774--1783, 2018.

\bibitem{passgoodpool}
Kaushik Manchella, Marina Haliem, Vaneet Aggarwal, and Bharat Bhargava.
\newblock Passgoodpool: Joint passengers and goods fleet management with reinforcement learning aided pricing, matching, and route planning.
\newblock {\em IEEE Transactions on Intelligent Transportation Systems}, 23(4):3866--3877, 2021.

\bibitem{flexpool}
Kaushik Manchella, Abhishek~K Umrawal, and Vaneet Aggarwal.
\newblock Flexpool: A distributed model-free deep reinforcement learning algorithm for joint passengers and goods transportation.
\newblock {\em IEEE Transactions on Intelligent Transportation Systems}, 22(4):2035--2047, 2021.

\bibitem{mangiaracina2019innovative}
Riccardo Mangiaracina, Alessandro Perego, Arianna Seghezzi, and Angela Tumino.
\newblock Innovative solutions to increase last-mile delivery efficiency in b2c e-commerce: a literature review.
\newblock {\em International Journal of Physical Distribution \& Logistics Management}, 2019.

\bibitem{CensusInfo}
{National Archives}.
\newblock {Census Criteria and Information}.

\bibitem{OpenStreetMap}
{OpenStreetMap}.
\newblock {}.

\bibitem{OverpassAPI}
{Overpass API}.
\newblock {}.

\bibitem{pillac2013review}
Victor Pillac, Michel Gendreau, Christelle Gu{\'e}ret, and Andr{\'e}s~L Medaglia.
\newblock A review of dynamic vehicle routing problems.
\newblock {\em European Journal of Operational Research}, 225(1):1--11, 2013.

\bibitem{prasetyo2021factors}
Yogi~Tri Prasetyo, Hans Tanto, Martinus Mariyanto, Christopher Hanjaya, Michael~Nayat Young, Satria~Fadil Persada, Bobby~Ardiansyah Miraja, and Anak Agung Ngurah~Perwira Redi.
\newblock Factors affecting customer satisfaction and loyalty in online food delivery service during the covid-19 pandemic: Its relation with open innovation.
\newblock {\em Journal of Open Innovation: Technology, Market, and Complexity}, 7(1):76, 2021.

\bibitem{Statista}
{Statista}.
\newblock {Statista eServices Report 2020}.

\bibitem{StatistaTaxi}
{Statista}.
\newblock The statistics portal. (2019). monthly number of uber’ s active users worldwide from 2016 to 2018 (in millions).

\bibitem{sutton}
Richard~S Sutton and Andrew~G Barto.
\newblock {\em Reinforcement learning: An introduction}.
\newblock MIT press, 2018.

\bibitem{vakulenko2019service}
Yulia Vakulenko, Poja Shams, Daniel Hellstr{\"o}m, and Klas Hjort.
\newblock Service innovation in e-commerce last mile delivery: Mapping the e-customer journey.
\newblock {\em Journal of Business Research}, 101:461--468, 2019.

\bibitem{warshall1962theorem}
Stephen Warshall.
\newblock A theorem on boolean matrices.
\newblock {\em Journal of the ACM (JACM)}, 9(1):11--12, 1962.

\end{thebibliography}
\bibliographystyle{plain}

\end{document}